%% file: N2Varxiv.tex
\documentclass[10pt,twocolumn,letterpaper]{article}

\usepackage{tikz}
\usetikzlibrary{patterns}
\usetikzlibrary{backgrounds}
\usepackage{cvpr}
\usepackage{times}
\usepackage{epsfig}
\usepackage{graphicx}
\usepackage{amsmath}
\usepackage{amssymb}
\usepackage{paralist}
\usepackage[perc]{overpic}
\usepackage{caption}
\usepackage{pifont}
\usepackage{placeins}
\usepackage[usestackEOL]{stackengine}
\usepackage{contour}
\contourlength{0.08em}
\contournumber{32}
\usepackage{ textcomp }

\newcommand{\figNum}[2]{\textcolor{#1}{\fontfamily{pnc}\selectfont\textbf{\small #2}}}
\newcommand{\figOverText}[2]{\textcolor{#1}{\fontfamily{pnc}\selectfont\textbf{\scriptsize{#2}}}}

\newcommand{\sign}{\boldsymbol{s}}

\newcommand{\noise}{\boldsymbol{n}}
\newcommand{\noised}{{\noise'}}
\newcommand{\img}{\boldsymbol{x}}
\newcommand{\imgd}{{\img'}}

\newcommand{\rfi}{\img_{\mathrm{RF}(i)}}
\newcommand{\rfni}{\tilde{\img}_{\mathrm{RF}({i})}}
\newcommand{\param}{\boldsymbol{\theta}}
\newcommand{\numPix}{N}

\newcommand{\E}[1]{\mathbb{E}\left[ #1 \right]}

\newcommand{\CARE}{\mbox{\textsc{CARE}}\xspace}

\newcommand{\NoiseNoise}{\mbox{\textsc{Noise2Noise}}\xspace}
\newcommand{\NoiseVoid}{\mbox{\textsc{Noise2Void}}\xspace}

\newcommand{\NtoN}{\mbox{\textsc{N2N}}\xspace}
\newcommand{\NtoV}{\mbox{\textsc{N2V}}\xspace}
\newcommand{\argmin}[1]{\underset{#1}{\operatorname{arg}\,\operatorname{min}}\;}

\usepackage[pagebackref=true,breaklinks=true,letterpaper=true,colorlinks,bookmarks=false]{hyperref}

\cvprfinalcopy 

\def\cvprPaperID{5541} 

\ifcvprfinal\fi
\begin{document}
	
	\title{Noise2Void - Learning Denoising from Single Noisy Images}
	
	\author{Alexander Krull$^{1,2}$, Tim-Oliver Buchholz$^{2}$, Florian Jug\\ 
		$^1$  {\tt\small krull@mpi-cbg.de} \\ 
		$^2$  {\small Authors contributed equally} \\ \\
		MPI-CBG/PKS (CSBD), Dresden, Germany
	}
	
	\maketitle
	\thispagestyle{empty}

\input{mainText.tex}

	{\small
		\bibliographystyle{ieee}
		\bibliography{bib}
	}
	
\end{document}

%% file: mainText.tex
\input{figures.tex}

\begin{abstract}
	The field of image denoising is currently dominated by discriminative deep 
	learning methods that are trained on pairs of noisy input and clean target 
	images.	Recently it has been shown that such methods can also be trained 
	without clean targets. Instead, independent pairs of noisy images can be 
	used, in an approach known as \NoiseNoise~(\NtoN). 
	Here, we introduce \NoiseVoid~(\NtoV), a training scheme that takes this idea 
	one step further. It does not require noisy image pairs, nor clean target 
	images. Consequently, \NtoV allows us to train directly on the body of data 
	to be denoised and can therefore be applied when other methods cannot. 
	Especially interesting is the application to biomedical image data, where 
	the acquisition of training targets, clean or noisy, is frequently not 
	possible.
	We compare the performance of \NtoV to approaches that have either clean 
	target images and/or noisy image pairs available. Intuitively, \NtoV cannot 
	be expected to outperform methods that have more information available 
	during training. Still, we observe that the denoising performance of 
	\NoiseVoid drops in moderation and compares favorably to 
	training-free denoising methods.
\end{abstract}


\section{Introduction}
\label{sec:intro}
Image denoising is the task of inspecting a noisy image $\img=\sign+\noise$ in 
order to separate it into two components: its signal $\sign$ and the signal 
degrading noise $\noise$ we would like to remove.
Denoising methods typically rely on the assumption that pixel values in $\sign$ 
are not statistically independent. 
In other words, observing the image context of an unobserved pixel might very 
well allow us to make sensible predictions on the pixel intensity.

A large body of work (\eg~\cite{roth2005fields, tappen2007learning}) explicitly 
modeled these interdependencies via Markov Random Fields (MRFs).
In recent years, convolutional neural networks (CNNs) have been trained in 
various ways to predict pixel values from surrounding image patches, \ie 
from the \emph{receptive field} of that pixel
~\cite{Weigert2018,lefkimmiatis2018universal,zhang2018ffdnet,guo2018toward,Weigert2017,zhang2017beyond,tai2017memnet,mao2016image}.

Typically, such systems require training pairs $(\img^j,\sign^j)$ of noisy 
input images $\img^j$ and their respective clean target images $\sign^j$ 
(ground truth).
Network parameters are then tuned to minimize an adequately formulated error 
metric (loss) between network predictions and known ground truth.

\figTeaser

Whenever ground truth images are not available, these methods cannot be trained 
and are therefore rendered useless for the denoising task at hand. 
Recent work by Lehtinen~\etal~\cite{noise2noise} offers an elegant solution for 
this problem. 
Instead of training a CNN to map noisy inputs to clean ground truth images, 
their \NoiseNoise~(\NtoN) training attempts to learn a mapping between pairs of 
independently degraded versions of the same training image, \ie~$(\sign + 
\noise,\sign + \noised)$, that incorporate the same signal $\sign$, but 
independently drawn noise $\noise$ and $\noised$.
Naturally, a neural network cannot learn to perfectly predict one noisy image 
from another one. 
However, networks trained on this impossible training task can produce results 
that converge to the same predictions as traditionally trained networks that do 
have access to ground truth images~\cite{noise2noise}.
In cases where ground truth data is physically unobtainable, \NtoN can still 
enable the training of denoising networks. 
However, this requires that two images capturing the same content ($\sign$) with
independent noises ($\noise, \noised$) can be acquired~\cite{buchholz2018cryo}.

Despite these advantages of \NtoN training, there are at least two shortcomings 
to this approach:
$(i)$~\NtoN training requires the availability of pairs of noisy images, and 
$(ii)$~the acquisition of such pairs with (quasi) constant $\sign$ is only possible 
for (quasi) static scenes.

Here we present \NoiseVoid~(\NtoV), a novel training scheme that overcomes both 
limitations.
Just as \NtoN, also \NtoV leverages on the observation that high quality 
denoising models can be trained without the availability of clean ground truth 
data.
However, unlike \NtoN or traditional training, \NtoV can also be applied to 
data for which neither noisy image pairs nor clean target images are available, \ie~\NtoV is a self-supervised training method.
In this work we make two simple statistical assumptions:
$(i)$~the signal $\sign$ is not pixel-wise independent, 
$(ii)$~the noise $\noise$ is conditionally pixel-wise independent given the signal~$\sign$.

We evaluate the performance of \NtoV on the BSD68 dataset \cite{roth2009fields}
and simulated microscopy data\footnote{For simulated microscopy data we know the perfect ground truth.}. 
We then compare our results to the ones obtained by a traditionally trained 
network~\cite{Weigert2018}, a \NtoN trained network and several self-supervised methods like 
BM3D~\cite{dabov2007image}, non-local means~\cite{buades2005non}, and to mean- and median-filters.
While it cannot be expected that our approach outperforms methods that have 
additional information available during training, we observe that the denoising 
performance of our results only drops moderately and is still outperforming 
BM3D.

Additionally, we apply \NtoV training and prediction to three biomedical 
datasets: cryo-TEM images from~\cite{buchholz2018cryo}, and two datasets from the Cell 
Tracking Challenge\footnote{http://celltrackingchallenge.net/}~\cite{ulman2017objective}.
For all these examples, the traditional training scheme cannot be applied due to the lack of ground truth data and \NtoN training 
is only applicable on the cryo-TEM data. This demonstrates the tremendous 
practical utility of our method. 

In summary, our main contributions are:
\begin{compactitem} 
	\item Introduction of \NoiseVoid, a novel approach for training denoising CNNs that 
	    requires only a body of single, noisy images.
	\item Comparison of our \NtoV trained denoising results to results obtained with existing 
		CNN training schemes~\cite{Weigert2018,noise2noise,zhang2017beyond} and non-trained 
		methods~\cite{tai2017memnet, buades2005non}.
	\item A sound theoretical motivation for our approach as well as a detailed 
		description of an efficient implementation.
\end{compactitem}

The remaining manuscript is structured as follows:
Section~\ref{sec:relate_work} contains a brief overview of related work.
In Section~\ref{sec:methods}, we introduce the baseline methods we 
later compare our own results to. 
This is followed by a detailed description of our proposed method and its 
efficient implementation.
All experiments and their results are described in 
Section~\ref{sec:experiments}, and our findings are finally discussed 
in Section~\ref{sec:conclusion}.

\section{Related Work}
\label{sec:relate_work}
Below, we will discuss other methods that consider not the denoising task as mentioned above, but instead the more general task of image restoration.
This includes the removal of perturbations such as JPEG artifacts or blur.
With \NtoV we have to stick to the more narrow task of denoising, as we rely on the fact that multiple noisy observations can help us to retrieve the true signal~\cite{noise2noise}. This is not the case for general perturbations such as blur.

We see \NtoV at the intersection of multiple methodological categories.
We will briefly discuss the most relevant works in each of them.
Note that \NtoN is omitted here, as it has been discussed above.

In concurrent work \cite{batson2019noise2self}, Batson \etal also introduce a method for self-supervised training of neural networks and other systems that is based on the idea of removing parts of the input. They show that this scheme can not only be applied by removing pixels, but also groups of variables in general.

\subsection{Discriminative Deep Learning Methods}
Discriminative deep learning methods are trained offline, extracting information from ground truth annotated training sets before they are applied to test data.

In~\cite{jain2009natural}, Jain \etal first apply CNNs for the denoising task.
They introduce the basic setup that is still used by successful methods today:
Denoising is seen as a regression task and the CNN learns to minimize a loss calculated between its prediction and clean ground truth data.

In~\cite{zhang2017beyond}, Zhang \etal achieve state-of-the-art results, by introducing a very deep CNN architecture for denoising.
The approach is based on the idea of residual learning~\cite{he2016deep}.
Their CNN attempts to predict not the clean signal, but instead the noise at every pixel, allowing for the computation of the signal in a subsequent step.
This structure allows them to train a single CNN for denoising of images corrupted by a wide range of noise levels.
Their architecture completely dispenses with pooling layers.

At about the same time Mao \etal introduce a complementary very deep encoder-decoder-architecture~\cite{mao2016image} for the denoising task.
They too make use of residual learning, but do so by introducing symmetric skip connections between the corresponding encoding and decoding modules.
Just as~\cite{zhang2017beyond}, they are able to use a single network for various levels of noise.

In~\cite{tai2017memnet} Tai \etal use recurrent persistent memory units as part of their architecture, and further improve on previous methods.

Recently Weigert~\etal presented the \CARE software framework for image restoration in the context of fluorescence microscopy data~\cite{Weigert2018}.
They acquire their training data by recording pairs of low- and high-exposure-images.
This can be a difficult procedure since the biological sample must not move between exposures.
We use their implementation as starting point for our experiments,
including their specific \emph{U-Net}~\cite{ronneberger2015u} architecture.

Note that \NtoV could in principle be applied with any of the mentioned architectures.
However,~\cite{tai2017memnet} and~\cite{zhang2017beyond} present an interesting peculiarity in this respect, as their residual architecture requires knowledge of the noisy input at each pixel.
In \NtoV, this input is masked when the gradient is calculated (see Section~\ref{sec:methods}).

\subsection{Internal Statistics Methods} 
Internal Statistics Methods do not have to be trained on ground truth data beforehand.
Instead, they can be directly applied to a test image where they extract all required information~\cite{zontak2011internal}.
\NtoV can be seen as member of this category, as it enables training directly on a test image.

In \cite{buades2005non}, Buades \etal introduced \emph{non-local means}, a classic denoising approach.
Like \NtoV, this method predicts pixel values based on their noisy surroundings.

BM3D, introduced by Dabov \etal~\cite{dabov2007image}, is a classic internal statistics based method.
It is based on the idea, that natural images usually contain repeated patterns. 
BM3D performs denoising of an image by grouping similar patterns together and jointly filtering them.
The downside of this approach is the computational cost during test time.
In contrast, \NtoV requires extensive computation only during training.
Once a CNN is trained for a particular kind of  data, it can be applied efficiently to any number of additional images.

In~\cite{UlyanovVL18}, Ulyanov \etal show that the structure of CNNs, inherently \emph{resonates} with the distribution of natural images and can be utilized for image restoration without requiring additional training data.
They feed a random but constant input into a CNN and train it to approximate a single noisy image as output.
Ulyanov \etal find that when they interrupt the training process at the right moment before convergence, the network produces a  regularized denoised image as output.

\subsection{Generative Models}
In~\cite{chen2018image}, Chen \etal present an image restoration approach based on \emph{generative adversarial networks} (GANs).
The authors use unpaired training samples consisting of noisy and clean images.
The GAN-generator learns to generate noise and create pairs of corresponding clean and noisy images, which are in turn used as training data in a traditional supervised setup.
Unlike N2V, this approach requires clean images during training.

Finally, we want to mention the work by Van Den Oord \etal~\cite{pixRNN}.
They present a generative model that is not used for denoising, but in spirit similar to \NtoV.
Like \NtoV, Van Den Oord \etal train a neural network to predict an unseen pixel value based on its surroundings.
The network is then used to generate synthetic images.
However, while we train our network for a regression task, they predict a probability distribution for each pixel.
Another difference lies in the structure of the receptive fields. 
While Van Den Oord~\etal use an asymmetric structure that is shifted over the image, we always mask the central pixel in a square receptive field.

\section{Methods }
\label{sec:methods}

Here, we will begin by discussing our image formation model.
Then, we will give a short recap of the traditional CNN training and of the \NtoN method.
Finally, we will introduce \NtoV and its implementation.

\subsection{Image Formation}
We see the generation of an image $\img=\sign+\noise$ as a draw from the joint distribution
\begin{equation}
    p(\sign,\noise)=p(\sign)p(\noise|\sign).
\end{equation}
We assume $p(\sign)$ to be an arbitrary distribution satisfying
\begin{equation}
    p(\sign_i|\sign_j)\neq p(\sign_i),
    \label{eq:signal}
\end{equation}
for two pixels $i$ and $j$ within a certain radius of each other.
That is, the pixels $\sign_i$ of the signal are not statistically independent.
With respect to the noise $\noise$, we assume a conditional distribution of the form 
\begin{equation}
    p(\noise|\sign)=\prod_i p(\noise_i|\sign_i).
    \label{eq:noise}
\end{equation}
That is, pixels values $\noise_i$ of the noise are conditionally independent given the signal.
We furthermore assume the noise to be zero-mean 
\begin{equation}
    \E{\noise_i}=0,
    \label{eq:expected_noise}
\end{equation}
which leads to
\begin{equation}
    \E{\img_i}=\sign_i.
    \label{eq:expected}
\end{equation}
In other words, if we were to acquire multiple images with the same signal, but different realizations of noise and average them, the result would approach the true signal.
An example of this would be recording multiple photographs of a static scene using a fixed tripod-mounted camera.

\subsection{Traditional Supervised Training} 
We are now interested in training a CNN to implement a mapping from $\img$ to $\sign$.
We will assume a fully convolutional network (FCN) \cite{long2015fully}, taking one image as input and predicting another one as output.

Here we want to take a slightly different but equivalent view on such a network.
Every pixel prediction $\hat{\sign}_i$ in the output of the CNN is has a certain \emph{receptive~field} $\rfi$ of input pixels,~\ie the set of pixels that influence the pixel prediction.
A pixel's receptive field is usually a square patch around that pixel.

Based on this consideration, we can also see our CNN as a function that takes a patch $\rfi$ as input and outputs a prediction $\hat{\sign}_i$ for the single pixel $i$ located at the patch center.
Following this view, the denoising of an entire image can be achieved by extracting overlapping patches and feeding them to the network one by one.
Consequently, we can define the CNN as the function
\begin{equation}
    f(\rfi;\param)=\hat{\sign}_i,
    \label{eq:prediction}
\end{equation}
where $\param$ denotes the vector of CNN parameters we would like to train.

In traditional supervised training we are presented with a set of training pairs $(\img^j,\sign^j)$, each consisting of a noisy input image $\img^j$ and a clean ground truth target $\sign^j$.
By again applying our patch-based view of the CNN, we can see our training data as pairs $(\rfi^j,\sign_i^j)$.
Where $\rfi^j$ is a patch around pixel $i$, extracted from training input image $\img^j$, and  $\sign^j_i$ is the corresponding target pixel value, extracted from the ground truth image $\sign^j$ at same position.
We now use these pairs to tune the parameters $\param$ to minimize pixel-wise loss
\begin{equation}
    \argmin{\param} \sum_j \sum_i L\left(f(\rfi^j;\param) = \hat{\sign}_i^j ,\sign_i^j \right).
    \label{eq:loss_sum}
\end{equation}
Here we consider the standard MSE loss
\begin{equation}
    L\left(\hat{\sign}_i^j ,\sign_i^j \right)=(\hat{\sign}_i^j-{\sign}_i^j)^2.
\end{equation}

\subsection{Noise2Noise Training}
Now let us consider the training procedure according to~\cite{noise2noise}.
\NtoN allows us to cope without clean ground truth training data.
Instead we start out with noisy image pairs $(\img^j,\imgd^j)$, where
\begin{equation}
\img^j=\sign^j + \noise^j \mbox{~and~} \imgd^j=\sign^j + \noised^j,
\label{eq:nois2noisImgs}
\end{equation}
 that is the two training images are identical up to their noise components $\noise^j$ and $\noised^j$, which are, in our image generation model, just two independent samples from the same distribution (see Eq.~\ref{eq:noise}).

We can now again apply our patch-based perspective and view our training data as pairs $(\rfi^j,\imgd^j_i)$ consisting of a noisy input patch $\rfi^j$, extracted from $\img^j$, and a noisy target $\imgd^j_i$, taken from $\imgd^j$ at the position $i$. 
As in traditional training, we tune our parameters to minimize a loss, similar to Eq.~\ref{eq:loss_sum},
this time however using our noisy target $\imgd^j_i$ instead of the ground truth signal ${\sign}^j_i$.
Even though we are attempting to learn a mapping from a noisy input to a noisy target, the training will still converge to the correct solution.
The key to this phenomenon lies in the fact that the expected value of the noisy input is equal to the clean signal \cite{noise2noise} (see Eq.~\ref{eq:expected}).

\figBlindSpot
\subsection{Noise2Void Training}
Here, we go a step further.
We propose to derive both parts of our training sample, the input and the target, from a single noisy training image $\img^j$.
If we were to simply extract a patch as input and use its center pixel as target, our network would just learn the identity, by directly mapping the value at the center of the input patch to the output (see Figure~\ref{fig:blind_spot}~a).

To understand how training from single noisy images is possible nonetheless, let us assume that we use a network architecture with a special receptive field.
We assume the receptive field $\rfni$ of this network to have a blind-spot in its center.
The CNN prediction $\hat{\sign_i}$ for a pixel is affected by all input pixels in a square neighborhood except for the input pixel $\img_i$ at its very location.
We term this type of network \emph{blind-spot network} (see Figure~\ref{fig:blind_spot}~b).

A blind-spot network can be trained using any of the training schemes described above.
Like with a normal network, we can apply the traditional training or \NtoN, using a clean target, or a noisy target respectively.
The blind-spot network has a little bit less information available for its predictions, and we can expect its accuracy to be slightly impaired compared to a normal network.
Considering however that only one pixel out of the entire receptive field is removed, we can assume it to still perform reasonably well.

The essential advantage of the blind-spot architecture is its inability to learn the identity.
Let us consider why this is the case.
Since we assume the noise to be pixel-wise independent given the signal (see Eq.~\ref{eq:noise}), the neighboring pixels carry no information about the value of $\noise_i$.
It is thus impossible for the network to produce an estimate that is better than its \emph{a priori} expected value (see Eq.~\ref{eq:expected_noise}).

The signal however is assumed to contain statistical dependencies (see Eq.~\ref{eq:signal}).
As a result, the network can still estimate the signal $\sign_i$ of a pixel by looking at its surroundings.

Consequently, a blind-spot network allows us to extract the input patch and target value from the same noisy training image.
We can train it by minimizing the empirical risk
\begin{equation}
      \argmin{\param} \sum_j \sum_i L\left(f(\rfni^j;\param) ,\img_i^j \right).
    \label{eq:loss_sum_n2v} 
\end{equation}
Note that the target $\img_i^j$, is just as good as the \NtoN target
$\imgd_i^j$, which has to be extracted from a second noisy image.
This becomes clear when we consider Eqs.~\ref{eq:nois2noisImgs}~and~\ref{eq:noise}:
The two target values $\img_i^j$ and $\imgd_i^j$ have an equal signal $\sign_i^j$ and their noise components are just two independent samples from the same distribution $p(\noise_i|\sign_i^j)$.
\figIdea

We have seen that a blind-spot network can in principle be trained using only individual noisy training images.
However, implementing such a network that can still operate efficiently is not trivial.
We propose a masking scheme to avoid this problem and achieve the same properties with any standard CNN:
We replace the value in the center of each input patch with a randomly selected value form the surrounding area (see supplementary material for details).
This effectively erases the pixel's information and prevents the network from learning the identity.

\subsection{Implementation Details}
\label{sec:implementation}
If we implement the above training scheme naively, it is unfortunately still not very efficient:
We have to process an entire patch to calculate the gradients for a single output pixel.
To mitigate this issue, we use the following approximation technique:
Given a noisy training image ${\img}_i$, we randomly extract patches of size $64 \times 64$ pixels, which are bigger than our networks receptive field (see supplementary material for details).
Within each patch we randomly select $\numPix$ pixels, using stratified sampling to avoid clustering.
We then mask these pixels and use the original noisy input values as targets at their position (see Figure~\ref{fig:n2vIdea}).
Further details on the masking scheme can be found in the supplementary note.
We can now simultaneously calculate the gradients for all of them, while ignoring the rest of the predicted image.
This is achieved using the standard \emph{Keras} pipeline with a specialized loss function that is zero for all but the selected pixels.
We use the CSBDeep framework~\cite{Weigert2017} as basis for our implementation.
Following the standard CSBDeep setup, we use a U-Net~\cite{ronneberger2015u} architecture, to which we added batch normalization~\cite{ioffe2015batch} before each activation function.

\section{Experiments}
\label{sec:experiments}
\figResults
We evaluate \NoiseVoid on natural images, simulated biological image data, and 
acquired microscopy images. 
\NtoV results are then compared to results of traditional and \NoiseNoise 
training, as well as results of training-free denoising methods like BM3D, non-local means, and
mean- and median filters.
Please refer to the supplementary material for more details on all experiments. 

\subsection{Denoising of BSD68 Data}
\label{sec:experiment68}
For the evaluation on natural image data we follow the example 
of~\cite{zhang2017beyond} and take $400$ gray scale images with $180\times180$ 
pixels as our training dataset. 
For testing we use the gray scale version of the BSD68 dataset. 
Noisy versions of all images are generated by adding zero mean Gaussian noise 
with standard deviation $\sigma = 25$.
Furthermore, we used data augmentation on the training dataset. 
More precisely, we rotated each image three times by $90^\circ$ and also added
all mirrored versions.
During training we draw random $64\times64$ pixel patches from this augmented 
training dataset.  

The network architecture we use for all BSD68 experiments is a 
U-Net~\cite{ronneberger2015u} with depth $2$, kernel size $3$, batch 
normalization, and a linear activation function in the last layer.
The network has $96$ feature maps on the initial level, which get doubled 
while the network gets deeper. 
We use a  learning rate of $0.0004$ and the default CSBDeep learning rate schedule,
halving the learning rate when a plateau on the validation loss is detected.

We used batch size $128$ for traditional training and batch size $16$ for \NoiseNoise, where we found that a larger batch leads to slightly diminished results. 
For \NoiseVoid training we use a batch size of $128$ and simultaneously 
manipulate $\numPix=64$ pixels per input patch (see Section~\ref{sec:implementation}), 
as before with an initial learning rate of $0.0004$.

In the first row of Figure~\ref{fig:n2vResults}, we compare our results 
to the ones obtained by BM3D, traditional training, and \NoiseNoise 
training. 
We report the average PSNR numbers on each dataset.
As mentioned earlier, \NtoV is not expected to outperform 
other training methods, as it can utilize less information for its prediction.
Still, here we observe that the denoising 
performance of \NtoV drops moderately below the performance of 
BM3D (which is not the case for other data).

\subsection{Denoising of Simulated Microscopy Data}
The acquisition of close to ground truth quality microscopy data is either 
impossible or at the very least, difficult and expensive. 
Since we need ground truth data to compute desired PSNR values, we decided to 
use a simulated dataset for our second set of experiments.
To this end, we simulated membrane labeled cells \emph{epithelia} and mimicked the 
typical image degradation of fluorescence microscopy by first applying Poisson 
noise and then adding zero mean Gaussian noise. 
We used this simulation scheme to generate high-SNR ground truth images and 
two corresponding low-SNR input images.
This data enables us to perform traditional, \NtoN, as well as \NtoV training.
We used the same data augmentation scheme as described in 
Section~\ref{sec:experiment68}.

The network architecture we use for all experiments on simulated data is a 
U-Net~\cite{ronneberger2015u} of depth $2$, kernel size $5$, batch norm, 
$32$ initial feature maps, and a linear activation function in the last layer.
Traditional and \NoiseNoise training was performed with batch size $16$ and 
an initial learning rate of $0.0004$. 
The \NoiseVoid training was performed with a batch size of $128$.
We simultaneously manipulate $\numPix=64$ pixels per input patch (see 
Section~\ref{sec:implementation}).
We again use the standard CSBDeep learning rate schedule for all three training methods.

In the second row of Figure~\ref{fig:n2vResults} one can appreciate the 
denoising quality of \NoiseVoid training, which reaches virtually the same
quality as traditional and \NoiseNoise training. 
All trained networks clearly outperform the results obtained by BM3D.

\subsection{Denoising of Real Microscopy Data}
As mentioned in the previous section, ground truth quality microscopy data is 
typically not available.
Hence, we can no longer compute PSNR values.

The network architecture we use for all experiments on real microscopy data
is a U-Net~\cite{ronneberger2015u} of depth $2$, kernel size $3$, batch norm, 
$32$ initial feature maps, and a linear activation function in the last layer.
For an efficient training of \NoiseVoid we simultaneously manipulate $\numPix=64$ 
pixels per input patch (see Section~\ref{sec:implementation}).
We use a batch size of $128$ and a initial learning rate of $0.0004$. 
For all three tasks we extracted random patches of $64 \times 64$ pixels and 
augmented them as described in previous sections.

\subsubsection{Cryo-TEM Data}
In cryo-TEM, the acquisition of high-SNR images is not possible due to beam 
induced damage~\cite{knapek1980beam}. 
Buchholz~\etal show in~\cite{buchholz2018cryo} how \NoiseNoise training can 
be applied to data acquired with a direct electron detector. 
To enable a qualitative assessment, we applied \NtoV to the same data as
in~\cite{buchholz2018cryo}. 

In the third row of Figure~\ref{fig:n2vResults}, we show the raw image data, 
results obtained by BM3D, \NoiseNoise results of~\cite{buchholz2018cryo}, 
and our \NoiseVoid results. 
The runtime of both trained methods is roughly equal and about $25$ times 
faster then the one of BM3D.
For better orientation we marked some known structures in the shown cryo-TEM 
image (see figure caption for details).
Unlike BM3D, the \NtoV trained network is able to preserve these as good as the 
\NtoN baseline.

\subsubsection{Fluorescence Microscopy Data}
Finally, we tested \NoiseVoid on fluorescence microscopy data from
the Cell Tracking Challenge. 
More specifically, we used the datasets Fluo-C2DL-MSC (CTC-MSC) and Fluo-N2DH-GOWT1 (CTC-N2DH). 
As before, no ground truth images or second noisy images are available. 
Hence, only BM3D and \NtoV training can be applied to this data.

In the last two rows of Figure~\ref{fig:n2vResults}, we compare our results to BM3D.
In the absence of ground truth data, we can only judge the results visually.
We find that the \NtoV trained network gives subjectively smooth and appealing result, while requiring only a fraction of the BM3D runtime.
\figLimitationRare
\figLimitationPattern

\subsection{Errors and Limitations}
We want to start this section by showing extreme error cases of \NtoV trained 
network predictions on real images (for which our training method performs least 
convincing).
Figure~\ref{fig:limitationRare} shows the ground truth image, and prediction 
results of traditionally trained and \NtoV trained networks. 
While the upper row contains the image with the largest squared single pixel 
error, the lower row shows the image with the largest sum of squared pixel 
errors.

We see these errors as an excellent illustration, showing a limitation of the \NtoV method.
One of the underlying assumptions of \NtoV is the predictability of the signal $\sign$ (see Eq.~\ref{eq:signal}).
Both test images shown in Figure~\ref{fig:limitationRare} include high irregularities, that are difficult to predict.  
The more difficult it is to predict a pixel's signal from its surroundings the more errors are expected to appear in \NtoV predictions.
This is of course true for traditional training and \NtoN as well.
However, while these methods can utilize the value in the center pixel of the receptive field, this value is blocked for \NtoV.

In Figure~\ref{fig:limitationPattern}, we illustrate another limitation of our method.
\NtoV cannot distinguish between the signal and structured noise that violates the assumption of pixel-wise independence (see Eq.~\ref{eq:noise}).
We demonstrate this behaviour using artificially generated structured noise applied to an image.
The \NtoV trained CNN removes the unpredictable components of the noise, but reveals the hidden pattern.
Interestingly, we find the same phenomenon in real microscopy data from the Fluo-C2DL-MSC dataset.
Denoising with a \NtoV trained CNN reveals a systematic error of the imaging system, visible as a striped pattern.

\begin{figure}[t]
	\centering
	\hspace{-0.3cm}\includegraphics[height=5.7cm]{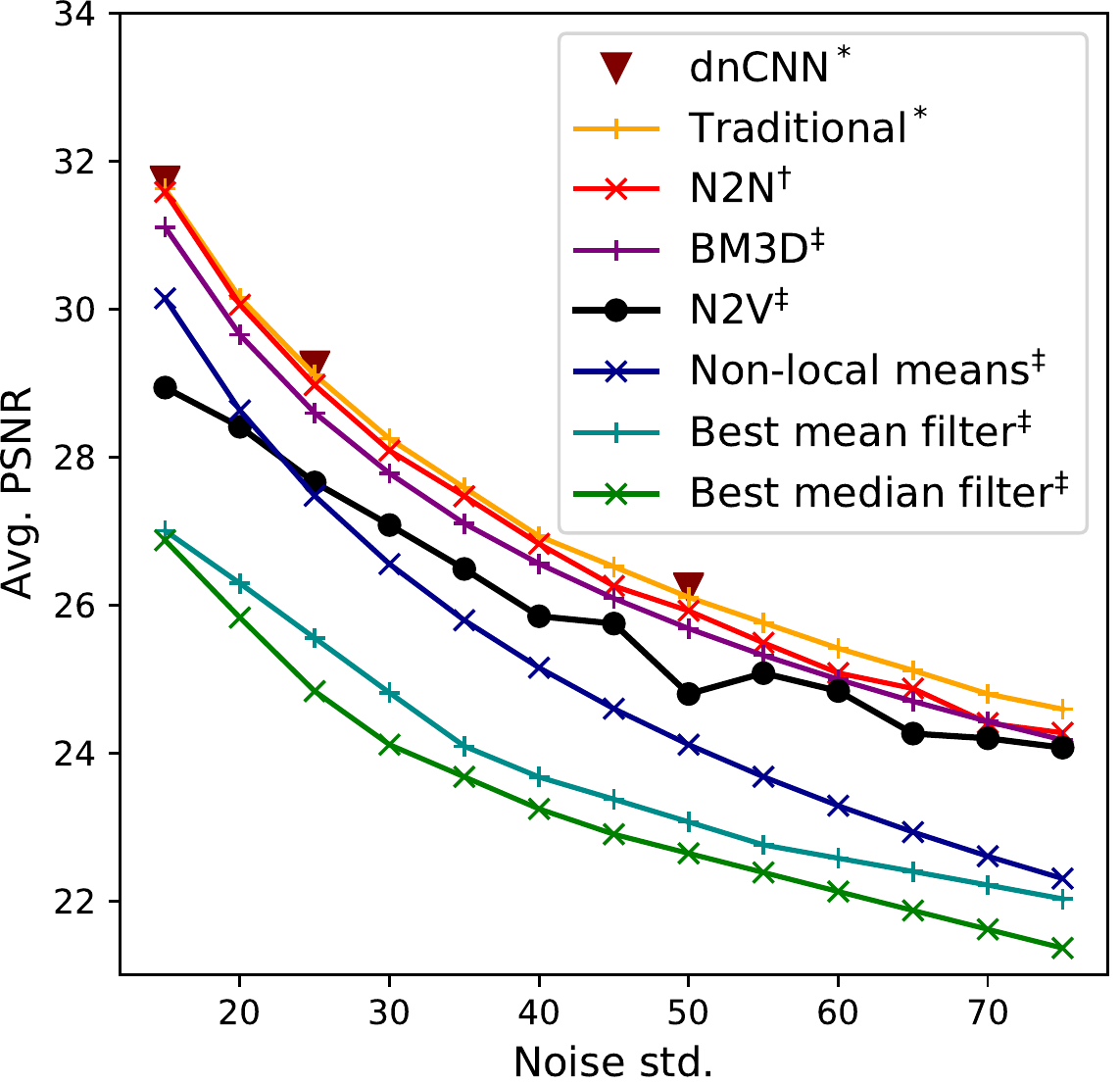}\hspace{-0.1cm}
	\includegraphics[height=5.7cm]{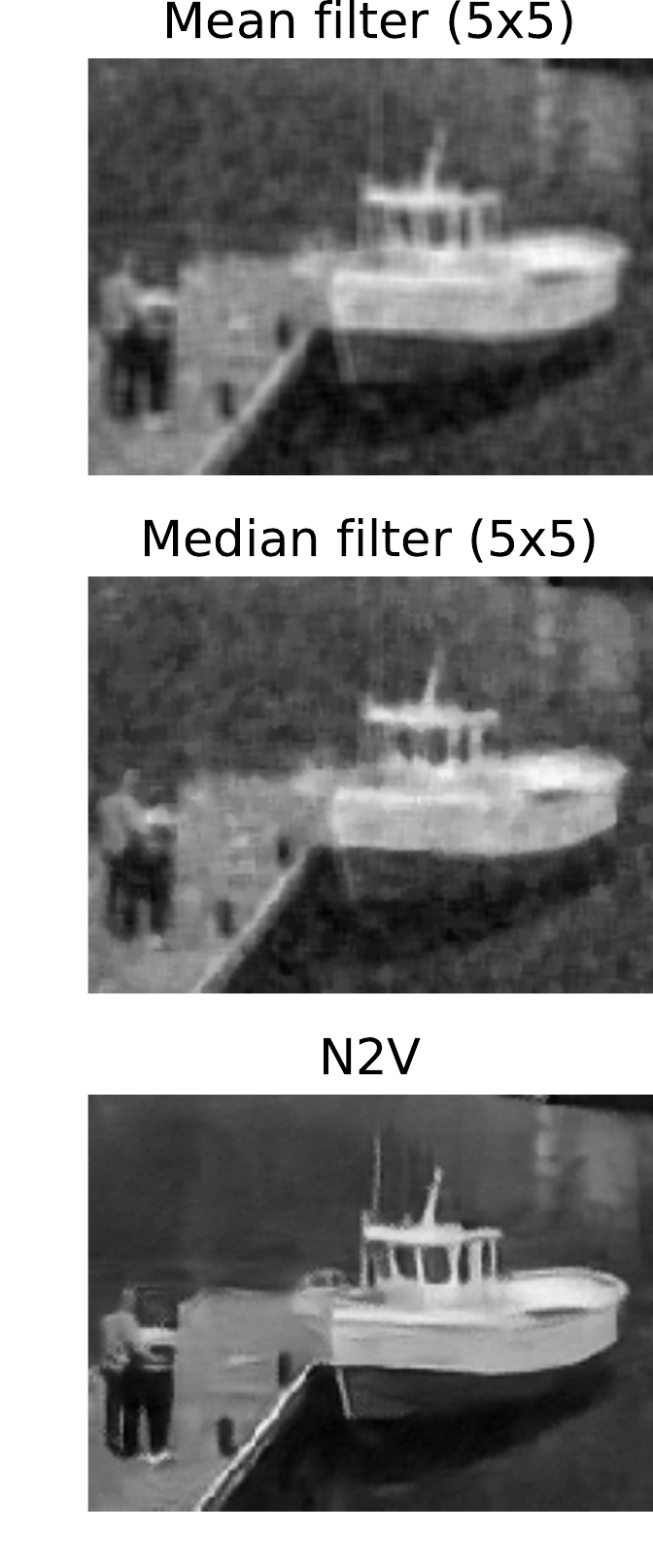}\hspace{-0.4cm}
	\caption{
		\label{fig:eval}
		Performance of \NtoV on the BSD68 dataset compared to various baselines.
		{\bf Left:} Average PSNR values as a function of the amount of added Gaussian noise.
		We consider square mean and median filters of 3, 5, and 7 pixels width/height, and show the best avg.~PSNR for each noise level.
		{$*$:} Method uses ground truth for training;
		{$\dagger$:} uses noisy image pairs;
		{$\ddagger$:} uses only single noisy images.
		{\bf Right:} Qualitative results of the best performing mean filer, median filter, and \NtoV on an image with Gaussian noise~(std.~$40$).
	}
	\vspace{-0.4cm}
\end{figure}

\subsection{Performance over Various Noise Levels}
We additionally ran our method and multiple baselines, including mean and median filters, as well as the classical non-local means \cite{buades2005non}, on the BSD68 dataset using various levels of noise.
To find the optimal parameter $h$ for non-local means we performed a grid search.
We also include a comparison to DnCNN using the numbers reported in \cite{zhang2017beyond}.
All results can be found in Figure~\ref{fig:eval}.

\section{Conclusion}
\label{sec:conclusion}
We have introduced \NoiseVoid, a novel training scheme that only requires single
noisy acquisitions to train denoising CNNs.
We have demonstrated the applicability of \NtoV on a variety of imaging modalities~\ie 
photography, fluorescence microscopy, and cryo-Transmission Electron Microscopy.
As long as our initial assumptions of a predictable signal and pixel-wise 
independent noise are met, \NtoV trained networks can compete with traditionally 
and \NtoN trained networks. 
Additionally, we have analyzed the behaviour of \NtoV training when these 
assumptions are violated.

We believe that the \NoiseVoid training scheme, as we propose it here, will allow
us to train powerful denoising networks. 
We have shown multiple examples how denoising networks can be trained on the same 
body of data which is to be processed in the first place.
Hence, \NtoV training will open the doors to a plethora of applications, \ie on 
biomedical image data.

\subsection*{Acknowledgements} 
We thank Uwe Schmidt, Martin Weigert, Alexander Dibrov, and Vladimir Ulman for the
 helpful discussions and for their assistance in data preparation.
We thank Tobias Pietzsch for proof reading.

%% file: figures.tex
\tikzset{
  schraffiert/.style={pattern=north east lines, pattern color=#1},
  schraffiert/.default=black
}

\newcommand\figIdea{
\begin{figure}[hbt]
\centerline{
\begin{minipage}{\linewidth}
\begin{tikzpicture}
\node[anchor=south west,inner sep=0] at (0,0) {
    \begin{overpic}[width=2.65cm]{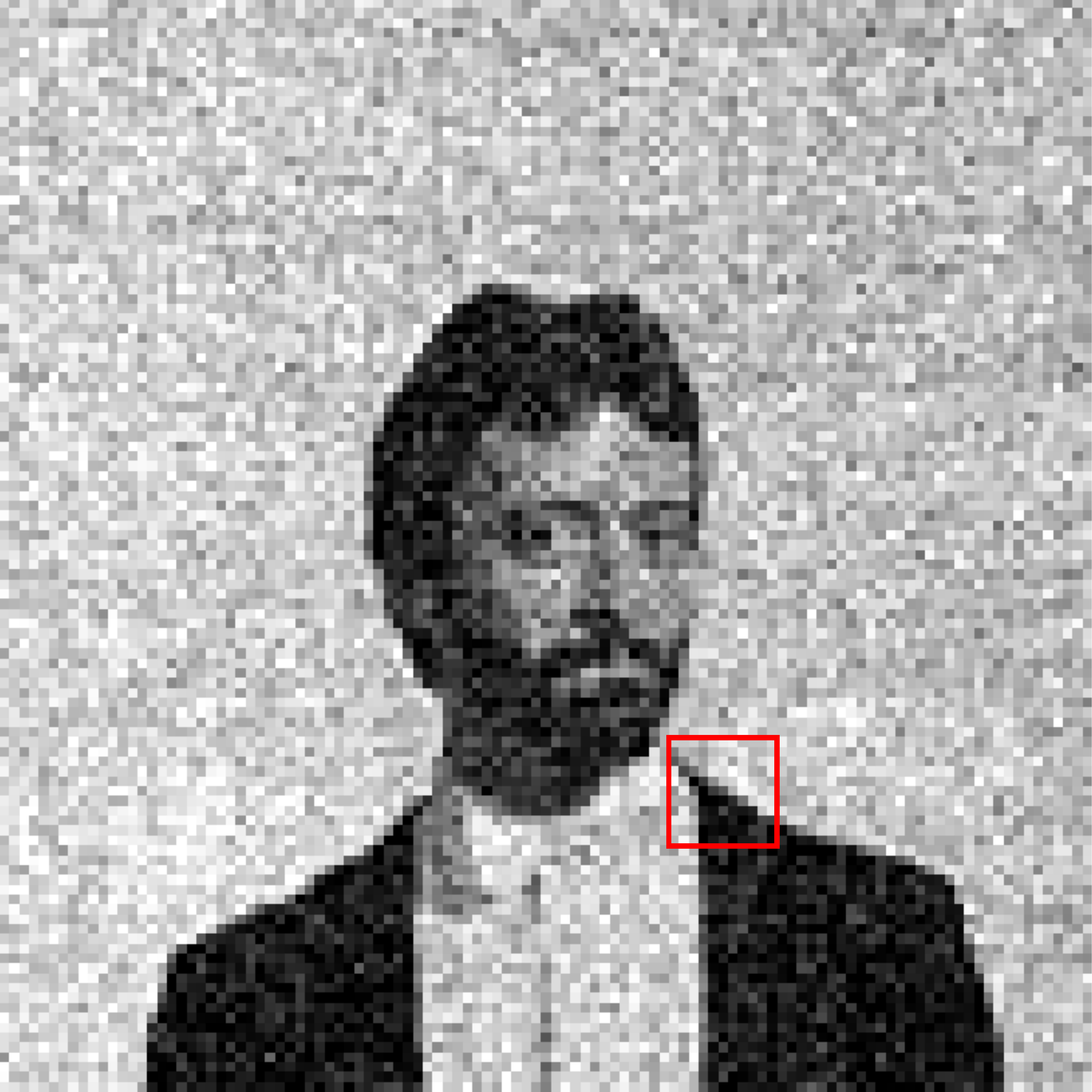}
    \end{overpic}};
\draw[red, dashed] (1.627,0.605) -- (2.815,0.012);
\draw[red, dashed] (1.629,0.87) -- (2.815,2.653);
\node[anchor=south west,inner sep=0] at (2.8,0) {
    \begin{overpic}[width=2.65cm]{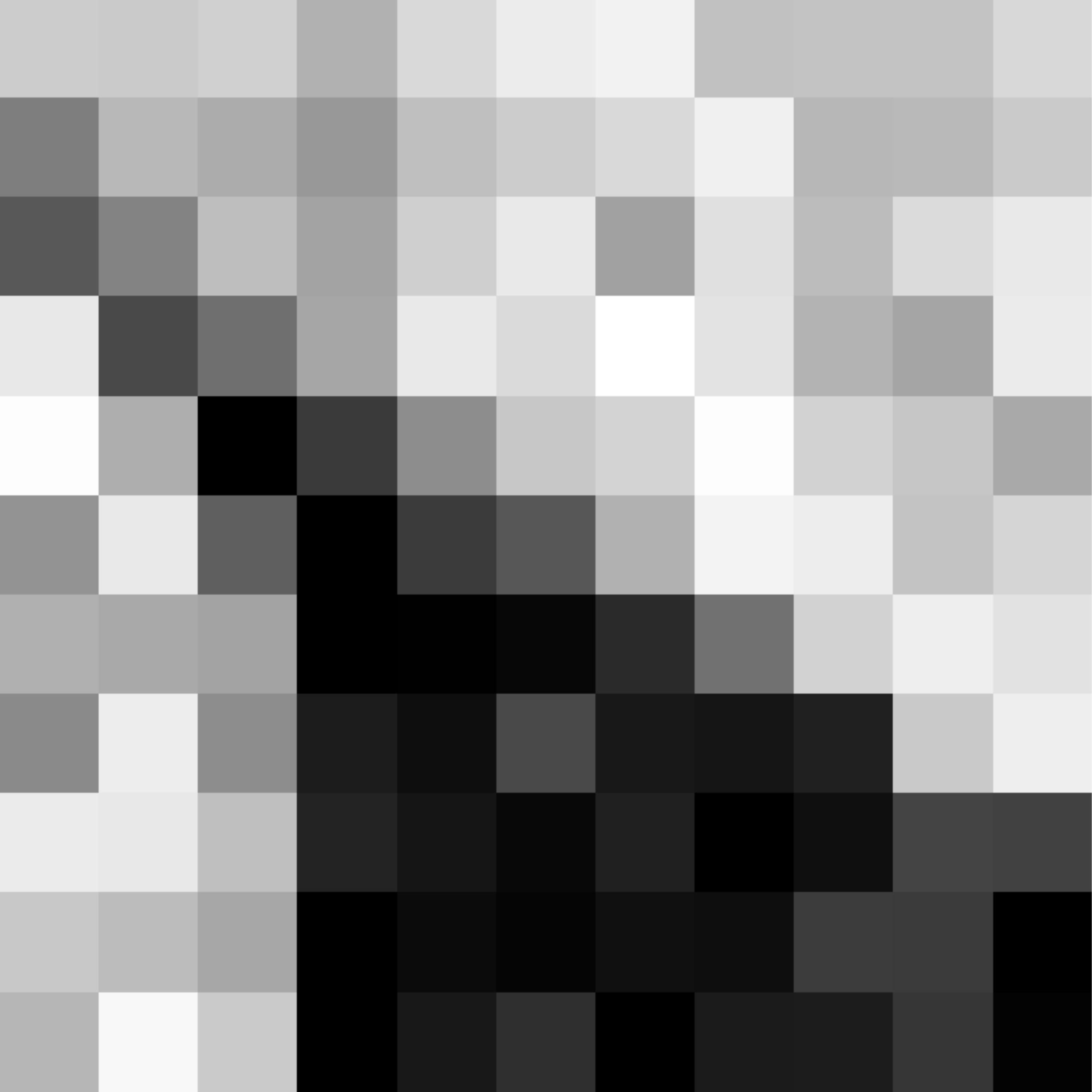}
    \end{overpic}};
\node[anchor=south west,inner sep=0] at (5.6,0) {
    \begin{overpic}[width=2.65cm]{figs/BSDS68_noisy0}
    \end{overpic}};
\def\centerpix{(4.01,1.21) rectangle (4.258,1.46)}
\draw[red, schraffiert=red]\centerpix;
\draw[red, schraffiert=red, semithick, even odd rule] (5.61,0.015) rectangle (8.25, 2.65) (6.81,1.21) rectangle (7.06,1.46);
\node[text width=0.5cm] at (0.3,0.2) {\textbf{(a)}};
\node[text width=0.5cm] at (3.1,0.2) {\textbf{(b)}};
\node[text width=0.5cm] at (5.9,0.2) {\textcolor{black}{\textbf{(c)}}};
\draw[->, blue, line width=0.2mm] (4.85,1.08) to[bend right] (4.25,1.33);
\draw[blue] (4.73,0.965) rectangle (4.978,1.215);
\end{tikzpicture}
\end{minipage}
}
\caption{Blind-spot masking scheme used during \NoiseVoid training.
{\bf (a)}~A noisy training image.
{\bf (b)}~A magnified image patch from (a).
During \NtoV training, a randomly selected pixel is chosen (blue rectangle) and its intensity copied over to create a blind-spot (red and striped square).
This modified image is then used as input image during training.
{\bf (c)}~The target patch corresponding to (b).
We use the original input with unmodified values also as target. The loss is only calculated for the blind-spot pixels we masked in (b).
}
\label{fig:n2vIdea}
\vspace{-2mm}
\end{figure}
}
\newcommand\figBlindSpot{
\begin{figure}[t]
    \centering
    \begin{overpic}[width=\linewidth]{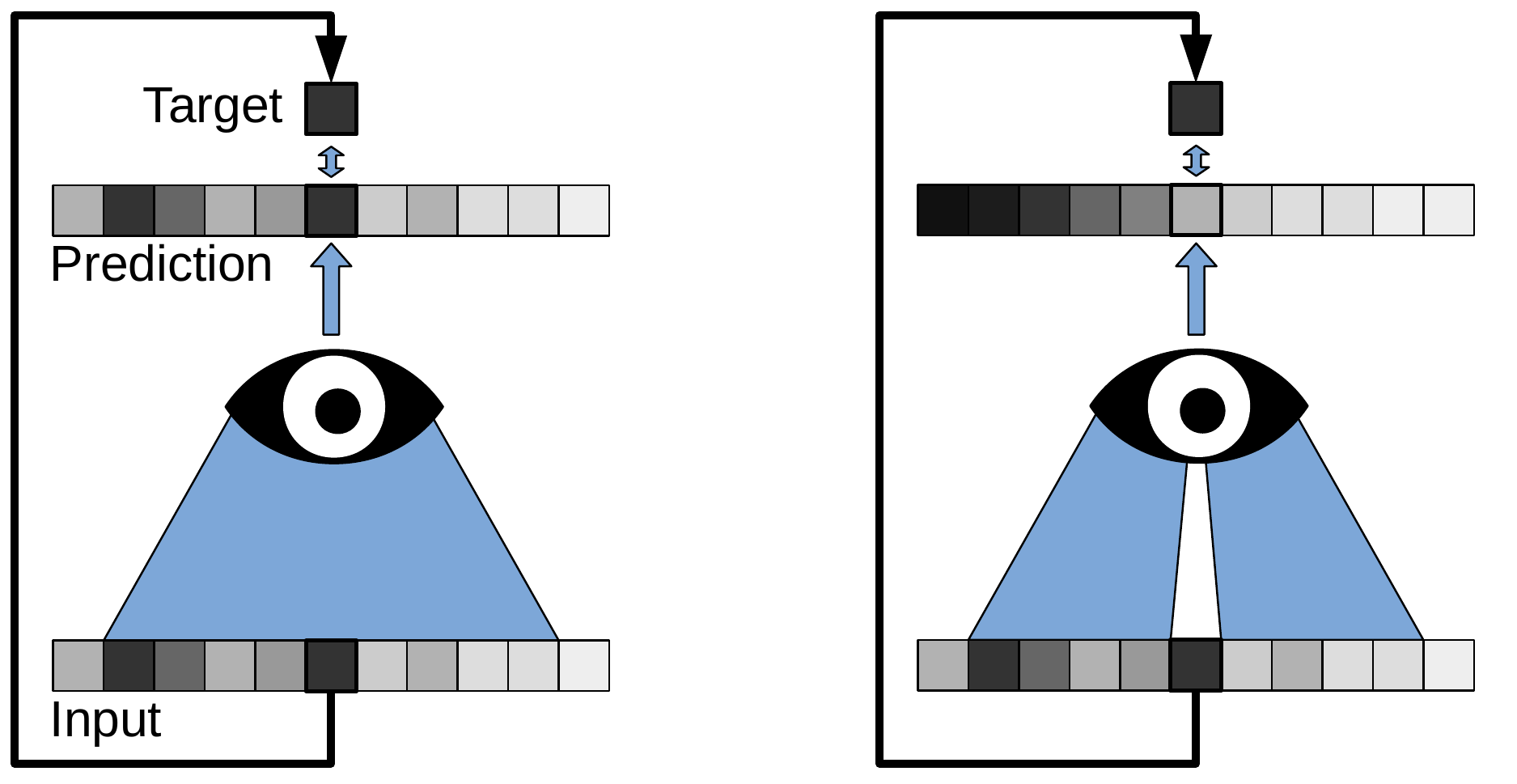}
        \put (1.5, -4) {\figNum{black}{(a)}}
        \put (59, -4) {\figNum{black}{(b)}}
    \end{overpic}\vspace{1em}
\caption{
A conventional network versus our proposed blind-spot network.
{\bf (a)}~In the conventional network the prediction for an individual pixel depends an a square patch of input pixels, known as a pixel's \emph{receptive field} (pixels under blue cone).
If we train such a network using the same noisy image as input and as target, the network will degenerate and simply learn the identity.
{\bf (b)}~In a \emph{blind-spot network}, as we propose it, the receptive field of each pixel excludes the pixel itself, preventing it from learning the identity.
We show that blind-spot networks can learn to remove pixel wise independent noise when they are trained on the same noisy images as input and target.
\label{fig:blind_spot}
}
\vspace{-2mm}
\end{figure}
}

\newcommand\figResults{
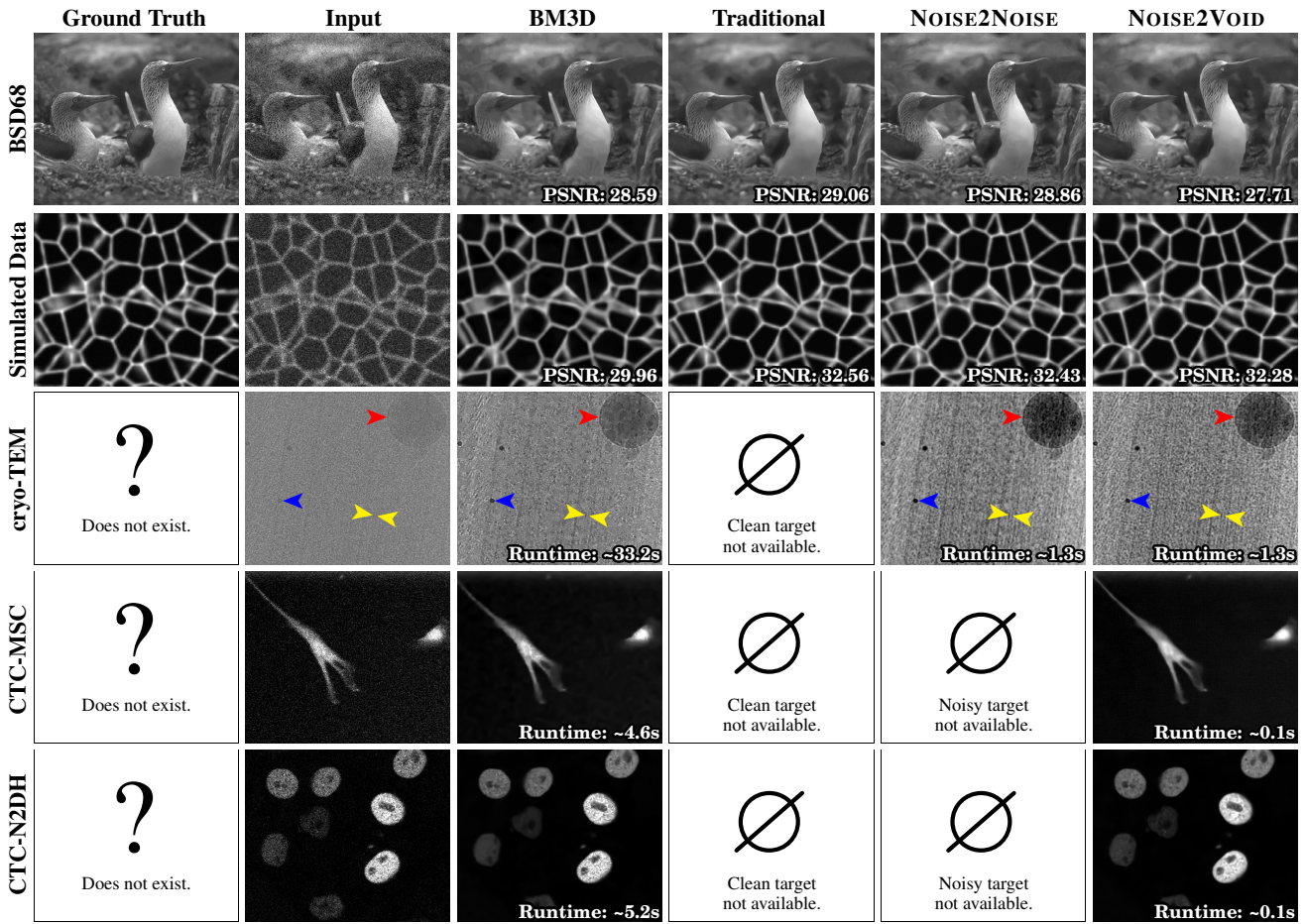
\begin{figure*}[t]
\vspace{-2mm}
\hspace{3.6mm}
\centerline{
\begin{minipage}{\linewidth}
\begin{tikzpicture}
\node[anchor=south west,inner sep=0] at (0,0) {
    {\begin{overpic}[width=2.74cm]{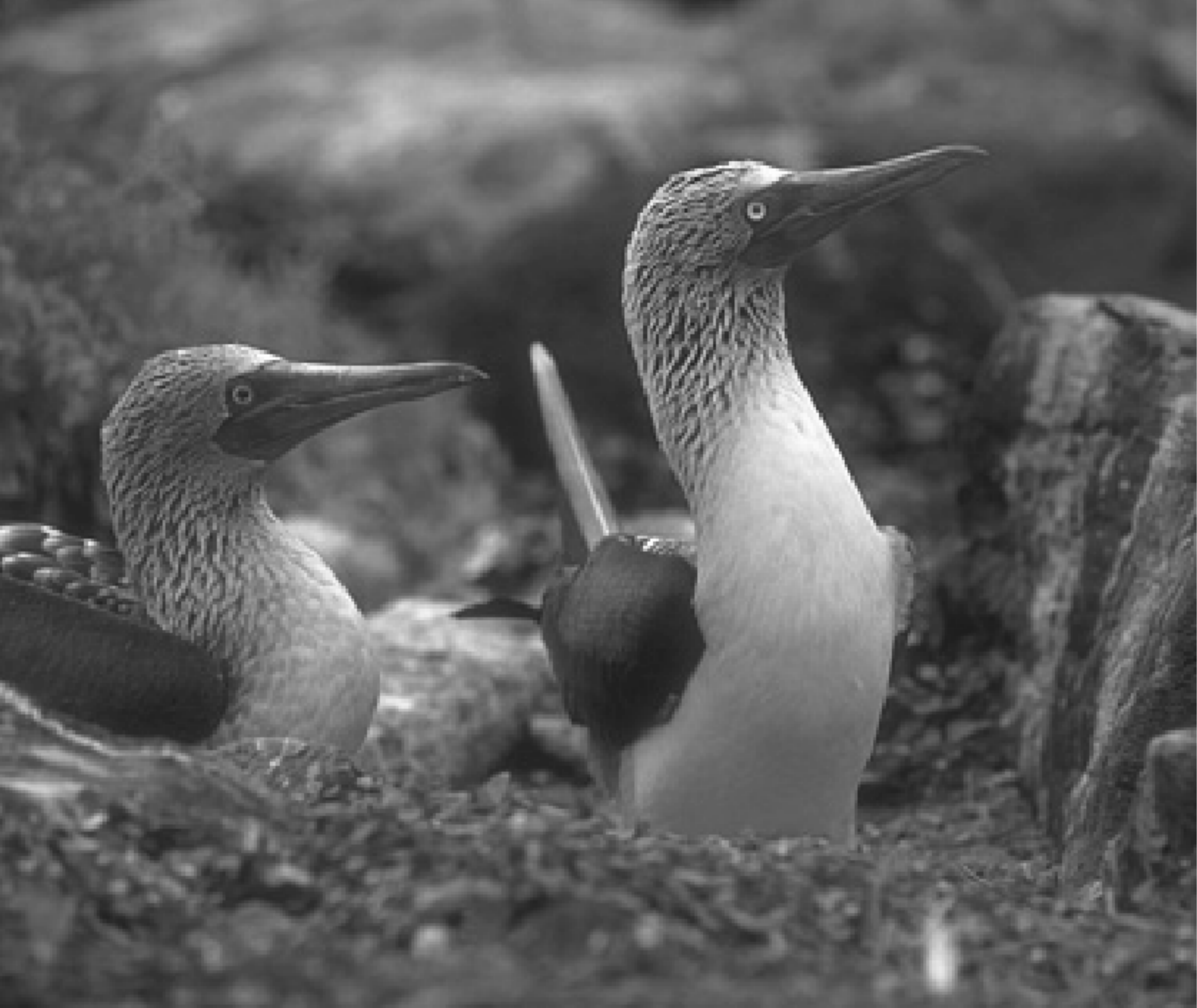}
        \put (14, 88) {\small\bf Ground Truth}
        \put (-12, 25) {\rotatebox{90}{\small\bf BSD68}}
    \end{overpic}}};
\node[anchor=south west,inner sep=0] at (2.84,0) {
    {\begin{overpic}[width=2.74cm]{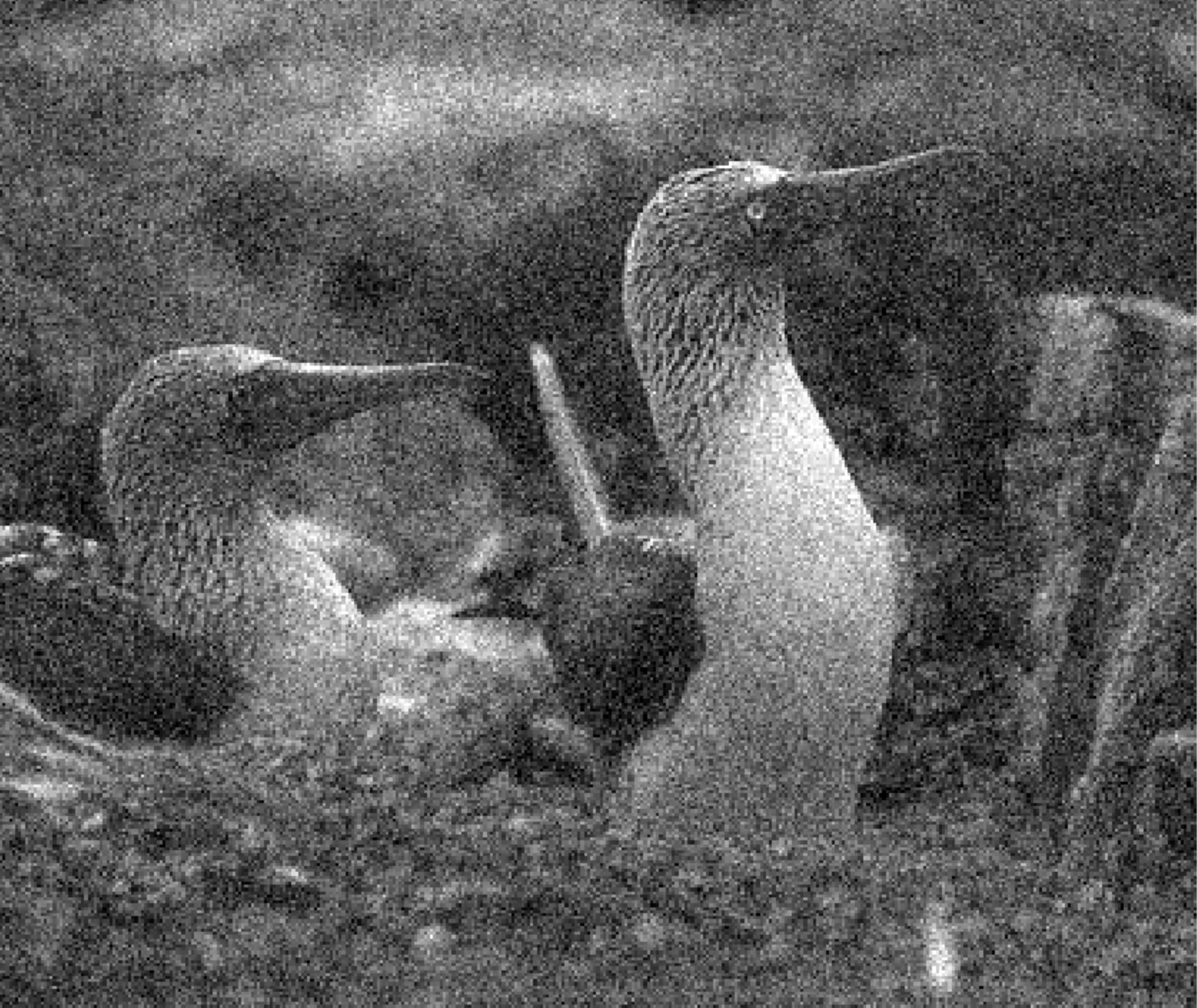}
        \put (39, 88) {\small\bf Input}
    \end{overpic}}};
\node[anchor=south west,inner sep=0] at (5.68,0) {
    {\begin{overpic}[width=2.74cm]{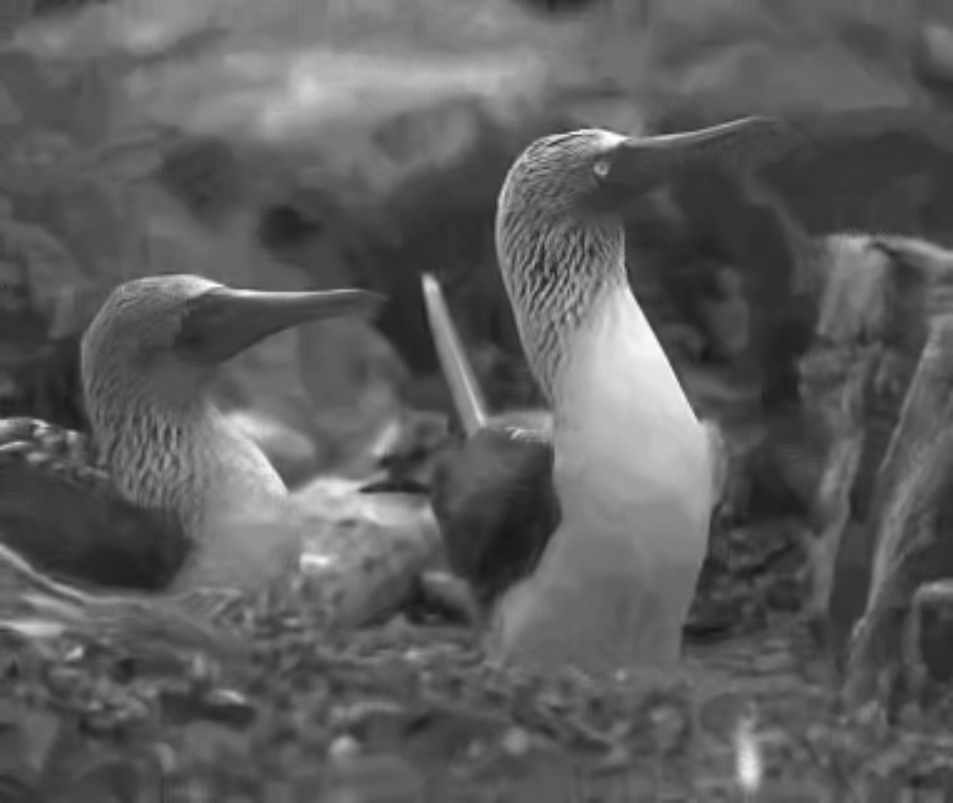}
        \put (35, 88) {\small\bf BM3D}
        \put (42, 2) {\contour{black}{\figOverText{white}{PSNR: 28.59}}}
    \end{overpic}}};
\node[anchor=south west,inner sep=0] at (8.52,0) {
    {\begin{overpic}[width=2.74cm]{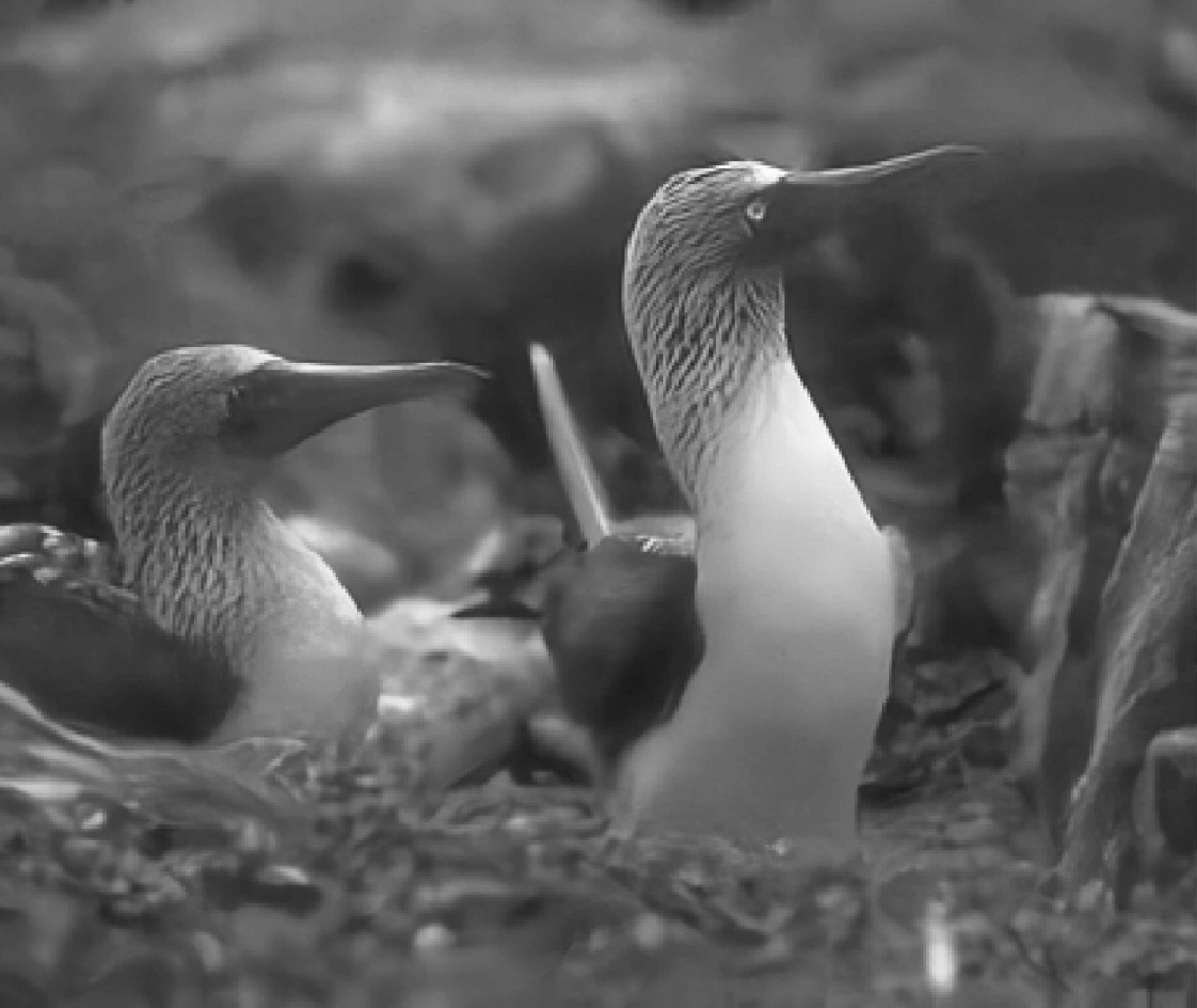}
        \put (21, 88) {\small\bf Traditional}
        \put (42, 2) {\contour{black}{\figOverText{white}{PSNR: 29.06}}}
    \end{overpic}}};
\node[anchor=south west,inner sep=0] at (11.36,0) {
    {\begin{overpic}[width=2.74cm]{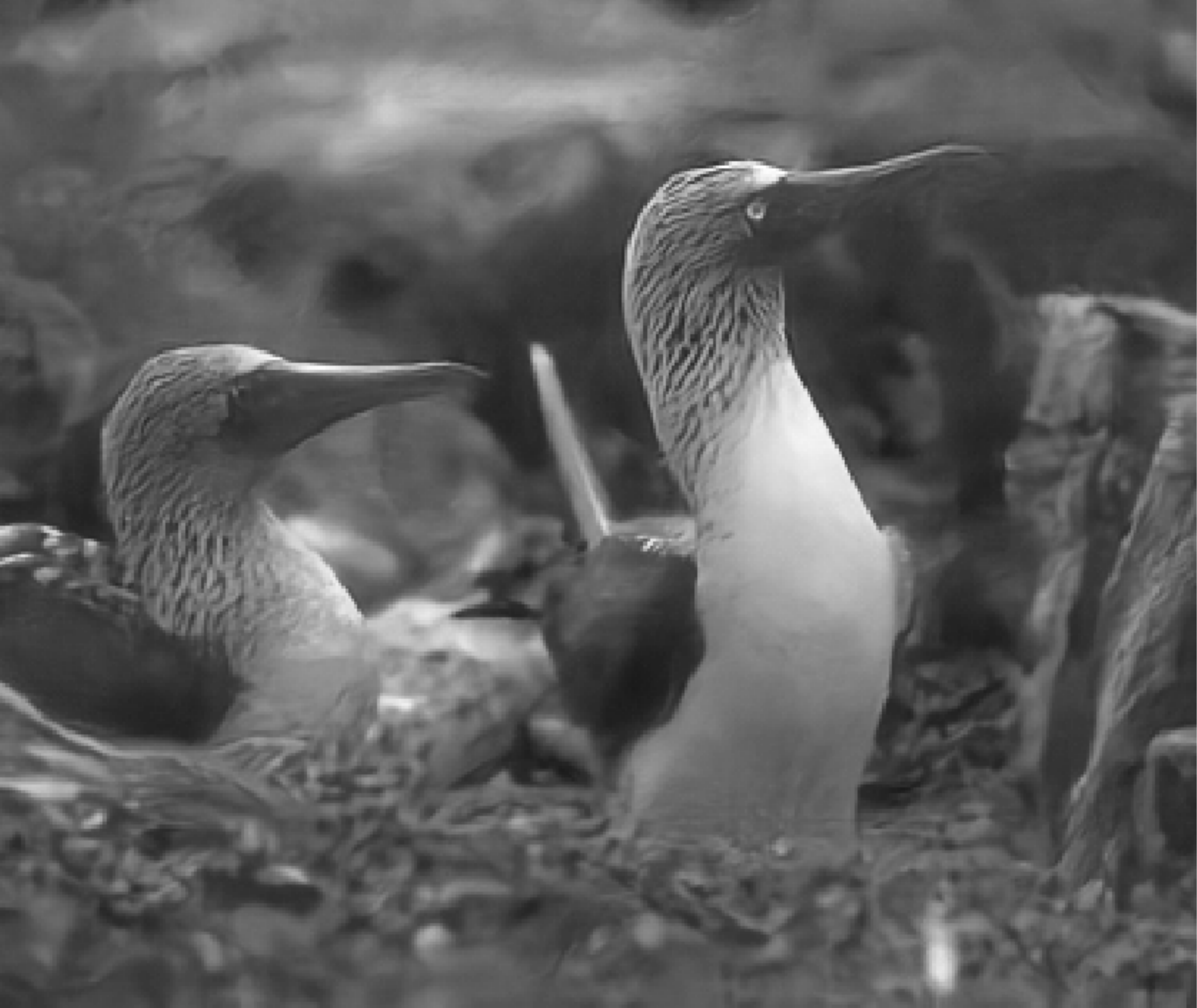}
        \put (14, 88) {\small\bf \NoiseNoise}
        \put (42, 2) {\contour{black}{\figOverText{white}{PSNR: 28.86}}}
    \end{overpic}}};
\node[anchor=south west,inner sep=0] at (14.20,0) {
    {\begin{overpic}[width=2.74cm]{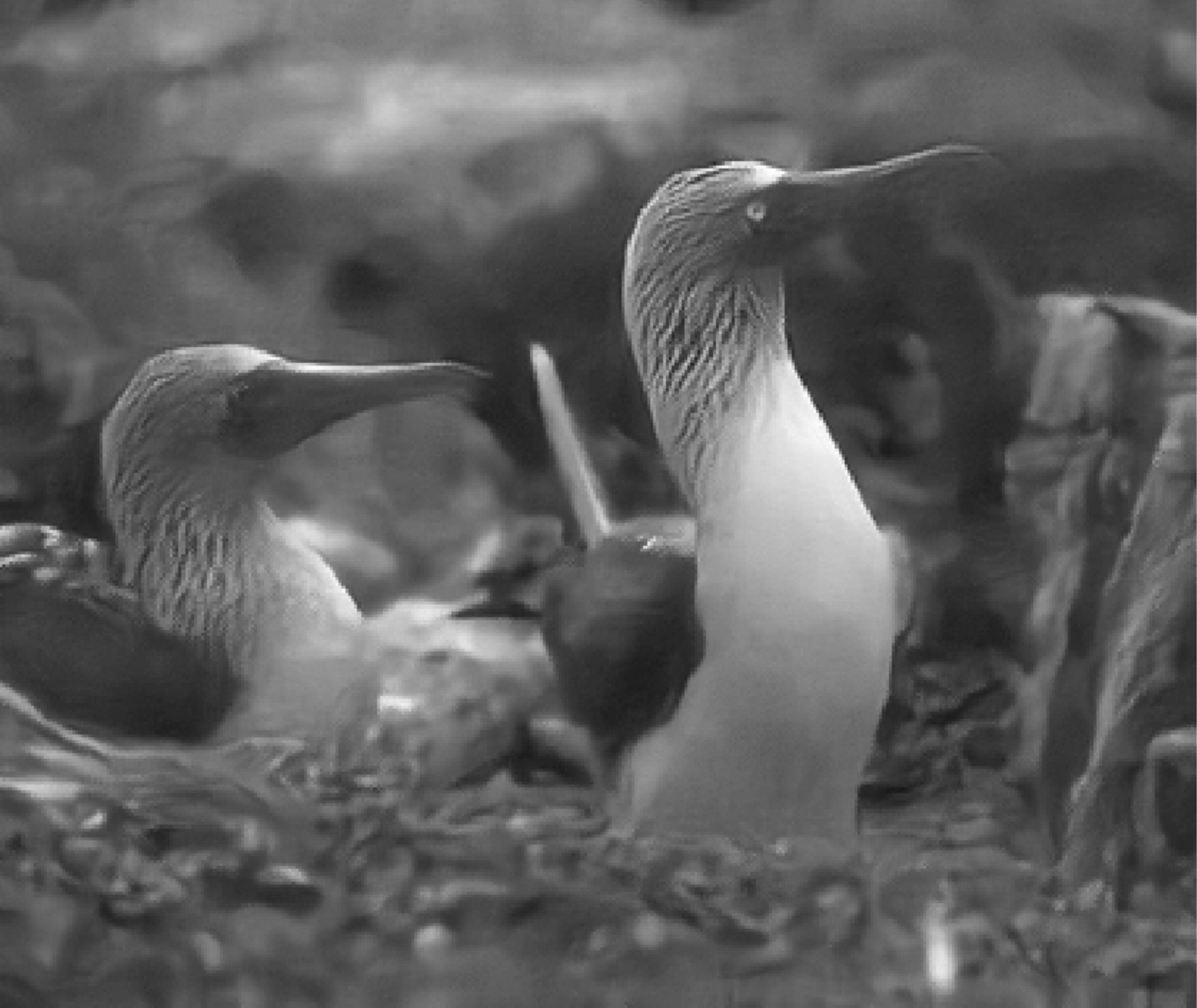}
        \put (17, 88) {\small\bf \NoiseVoid}
        \put (42, 2) {\contour{black}{\figOverText{white}{PSNR: 27.71}}}
    \end{overpic}}};
\end{tikzpicture}
\end{minipage}
}

\vspace{0.9mm}
\hspace{3.6mm}
\centerline{
\begin{minipage}{\linewidth}
\begin{tikzpicture}
\node[anchor=south west,inner sep=0] at (0,0) {
    {\begin{overpic}[width=2.74cm]{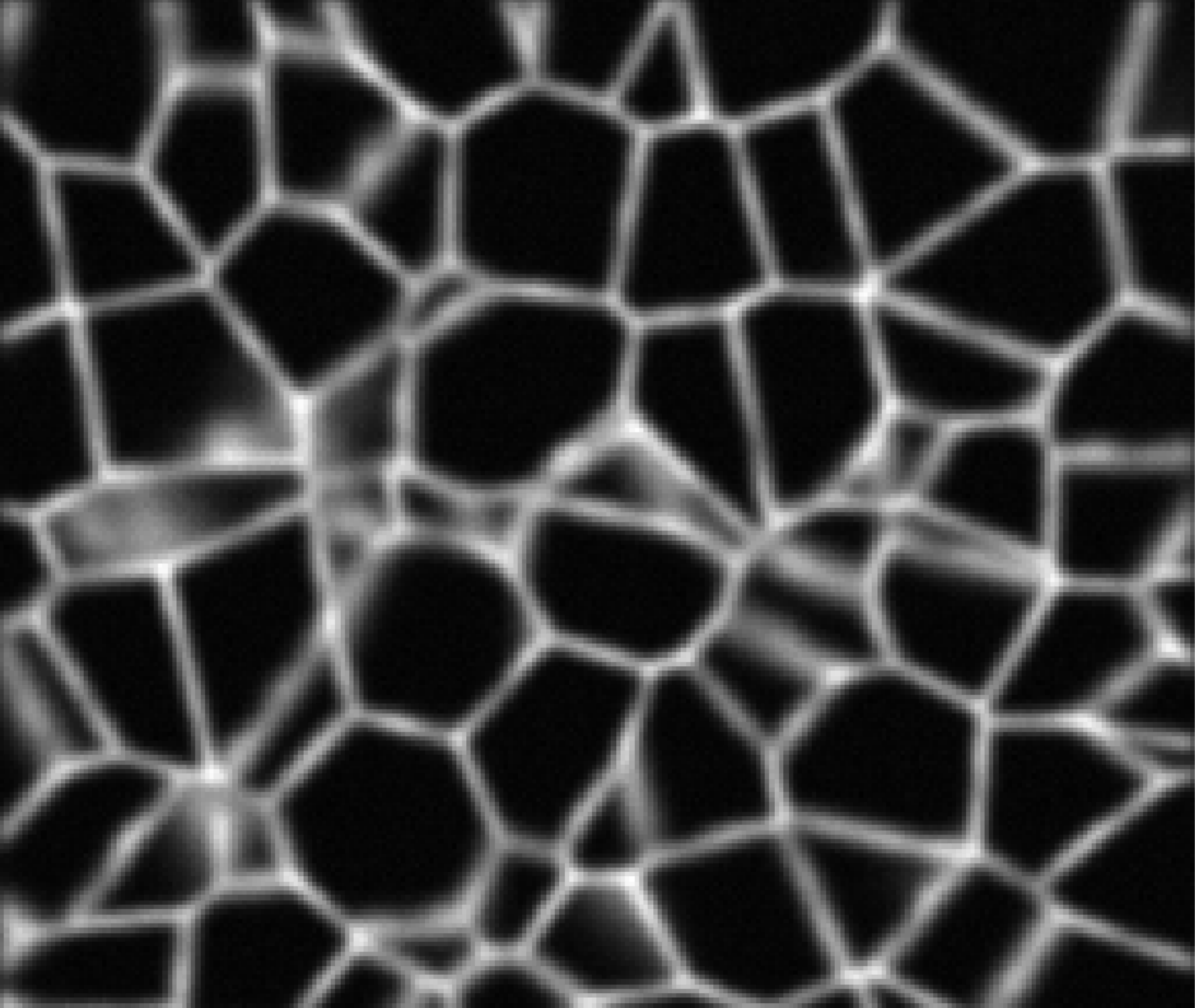}
        \put (-12, 4) {\rotatebox{90}{\small\bf Simulated Data}}
    \end{overpic}}};
\node[anchor=south west,inner sep=0] at (2.84,0) {
    {\begin{overpic}[width=2.74cm]{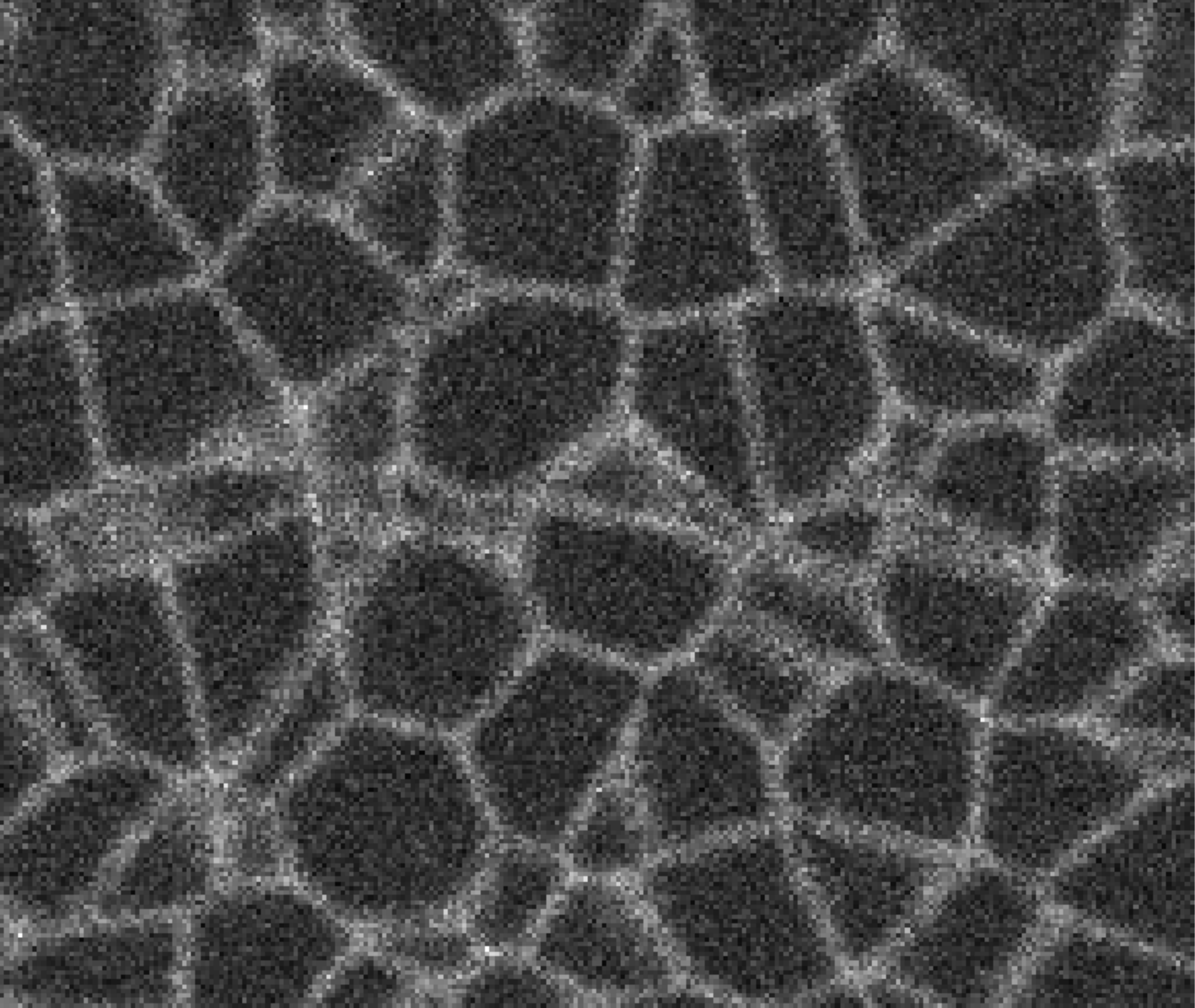}
    \end{overpic}}};
\node[anchor=south west,inner sep=0] at (5.68,0) {
    {\begin{overpic}[width=2.74cm]{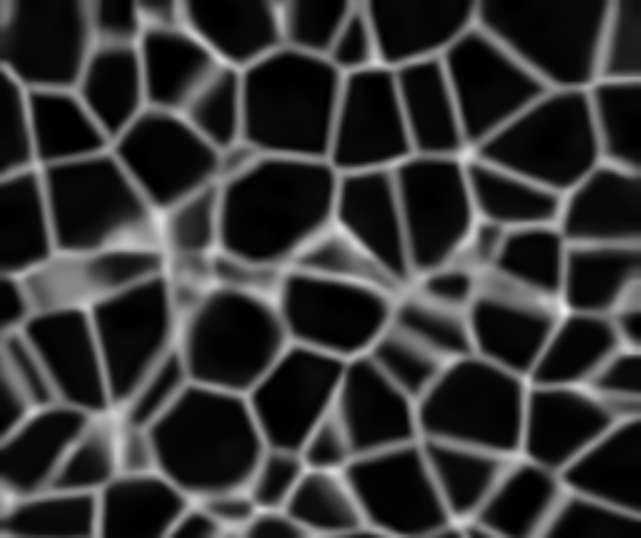}
        \put (42, 2) {\contour{black}{\figOverText{white}{PSNR: 29.96}}}
    \end{overpic}}};
\node[anchor=south west,inner sep=0] at (8.52,0) {
    {\begin{overpic}[width=2.74cm]{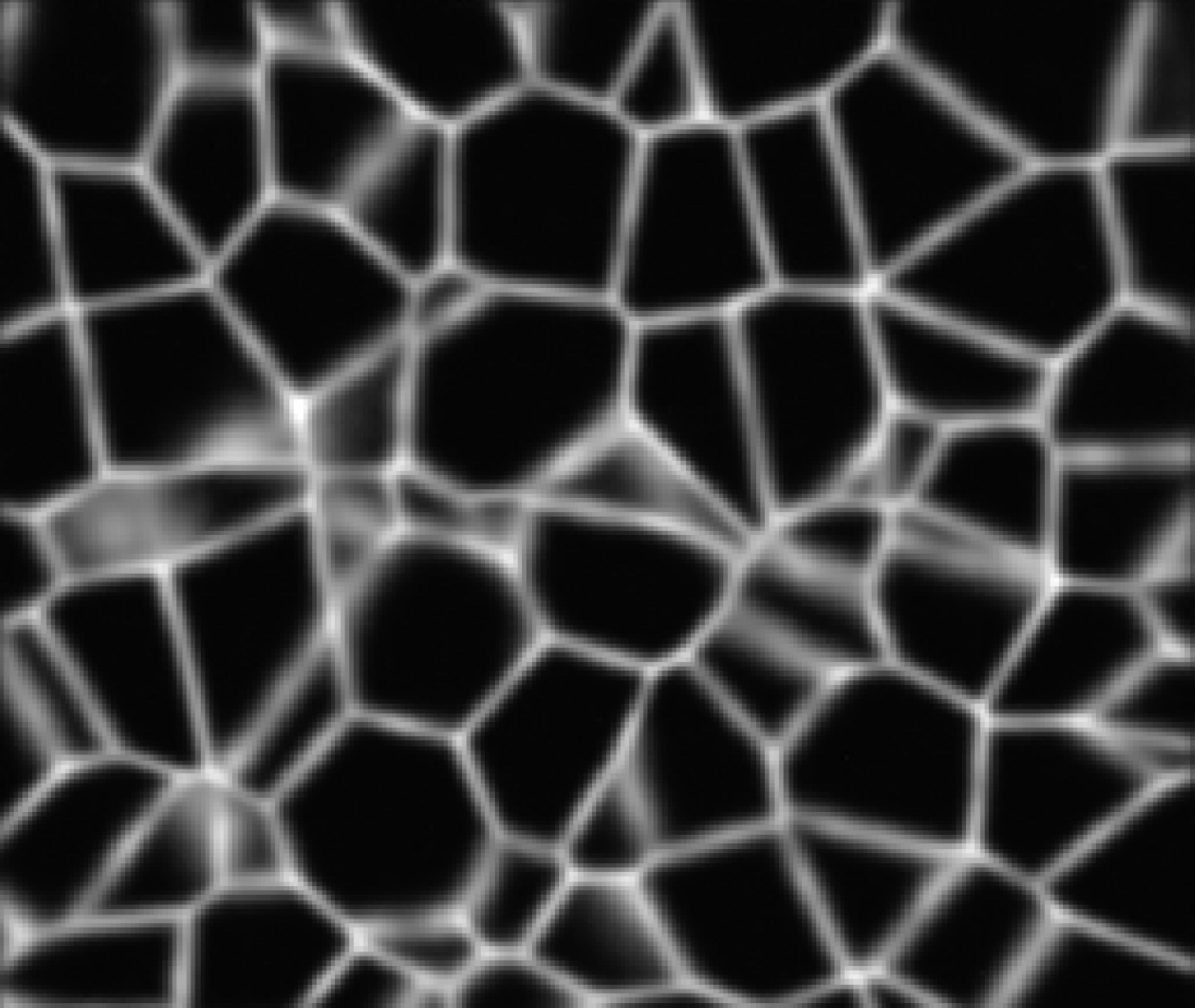}
        \put (42, 2) {\contour{black}{\figOverText{white}{PSNR: 32.56}}}
    \end{overpic}}};
\node[anchor=south west,inner sep=0] at (11.36,0) {
    {\begin{overpic}[width=2.74cm]{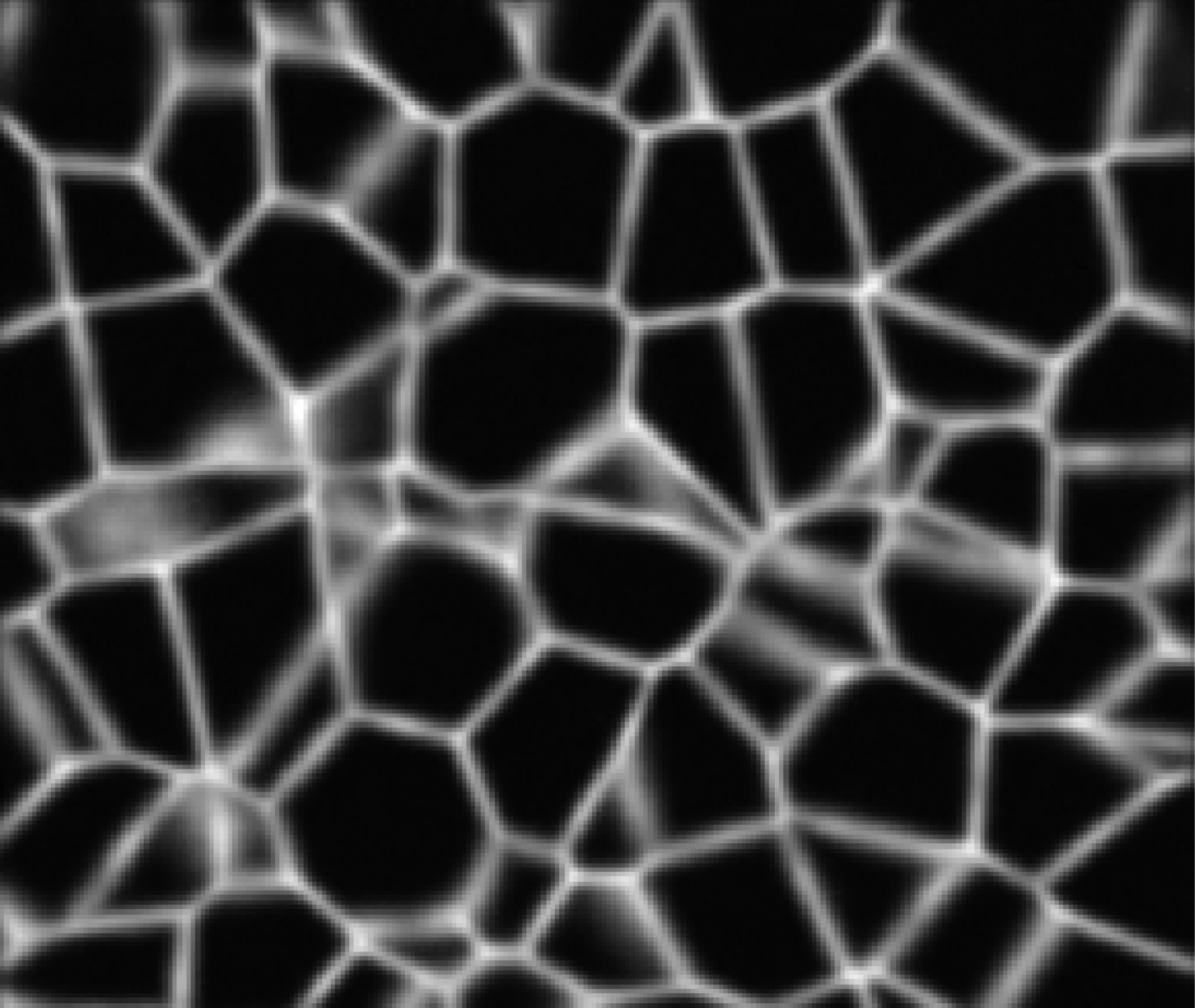}
        \put (42, 2) {\contour{black}{\figOverText{white}{PSNR: 32.43}}}
    \end{overpic}}};
\node[anchor=south west,inner sep=0] at (14.20,0) {
    {\begin{overpic}[width=2.74cm]{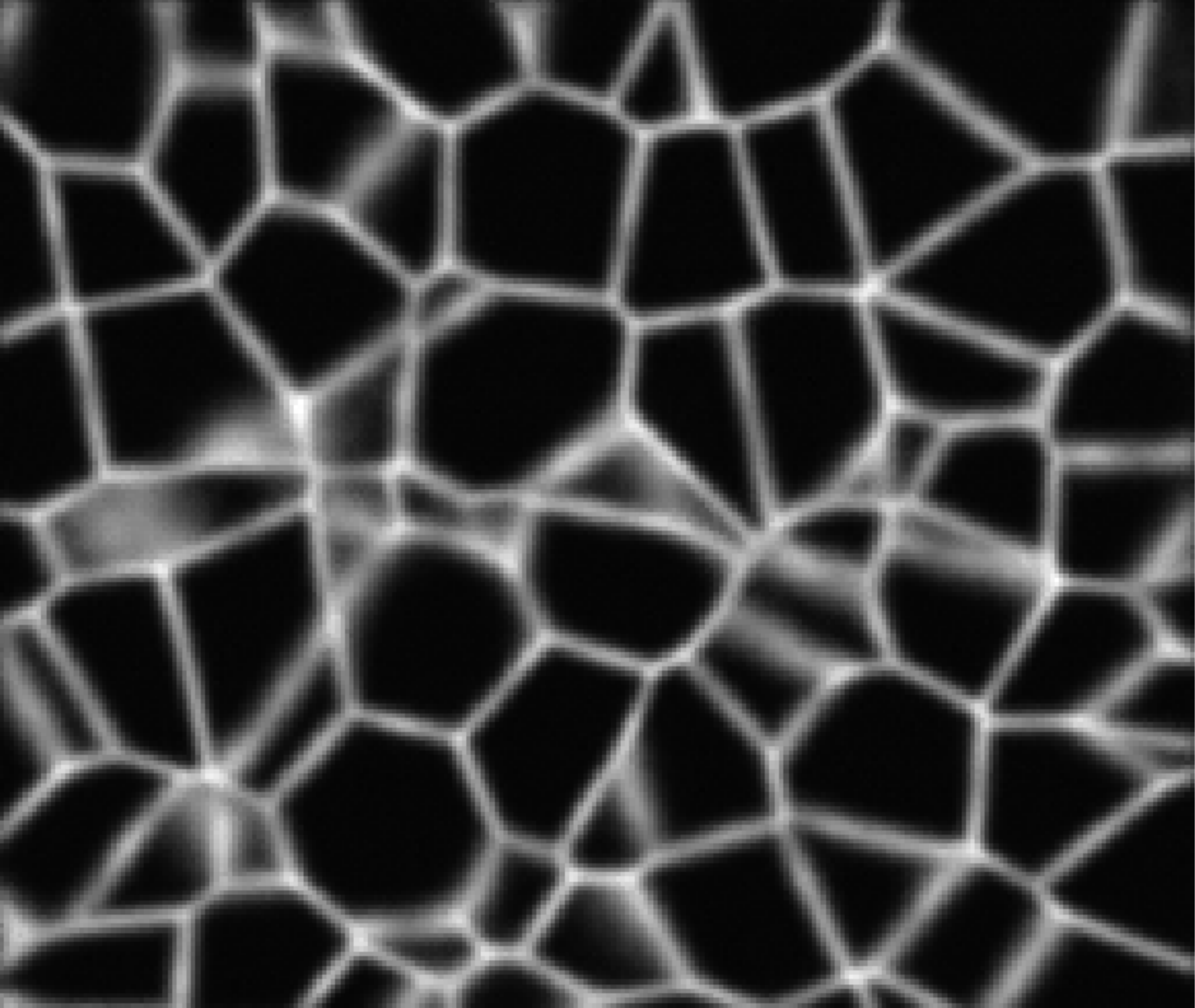}
        \put (42, 2) {\contour{black}{\figOverText{white}{PSNR: 32.28}}}
    \end{overpic}}};
\end{tikzpicture}
\end{minipage}
}

\vspace{0.6mm}
\hspace{3.6mm}
\centerline{
\begin{minipage}{\linewidth}
\begin{tikzpicture}
\node[anchor=south west,inner sep=0] at (0,0) {
    {\begin{overpic}[width=2.74cm]{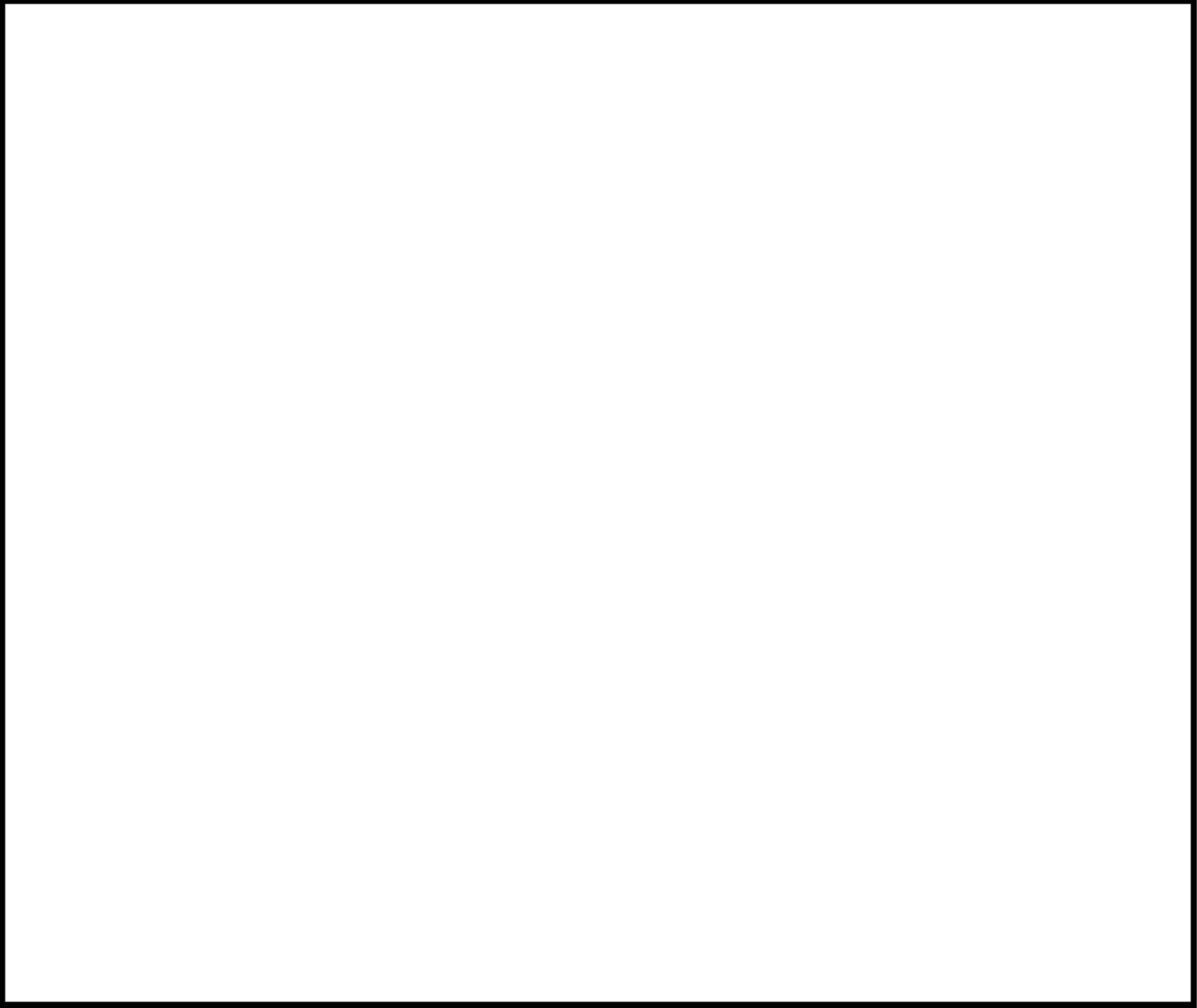}
        \put (37, 34) {\scalebox{4}{?}}
        \put (-1, 16) {\makebox[80pt]{\Centerstack{\scriptsize{Does not exist.}}}}
        \put (-12, 16) {\rotatebox{90}{\small\bf cryo-TEM}}
    \end{overpic}}};
\node[anchor=south west,inner sep=0] at (2.84,0) {
    {\begin{overpic}[width=2.74cm]{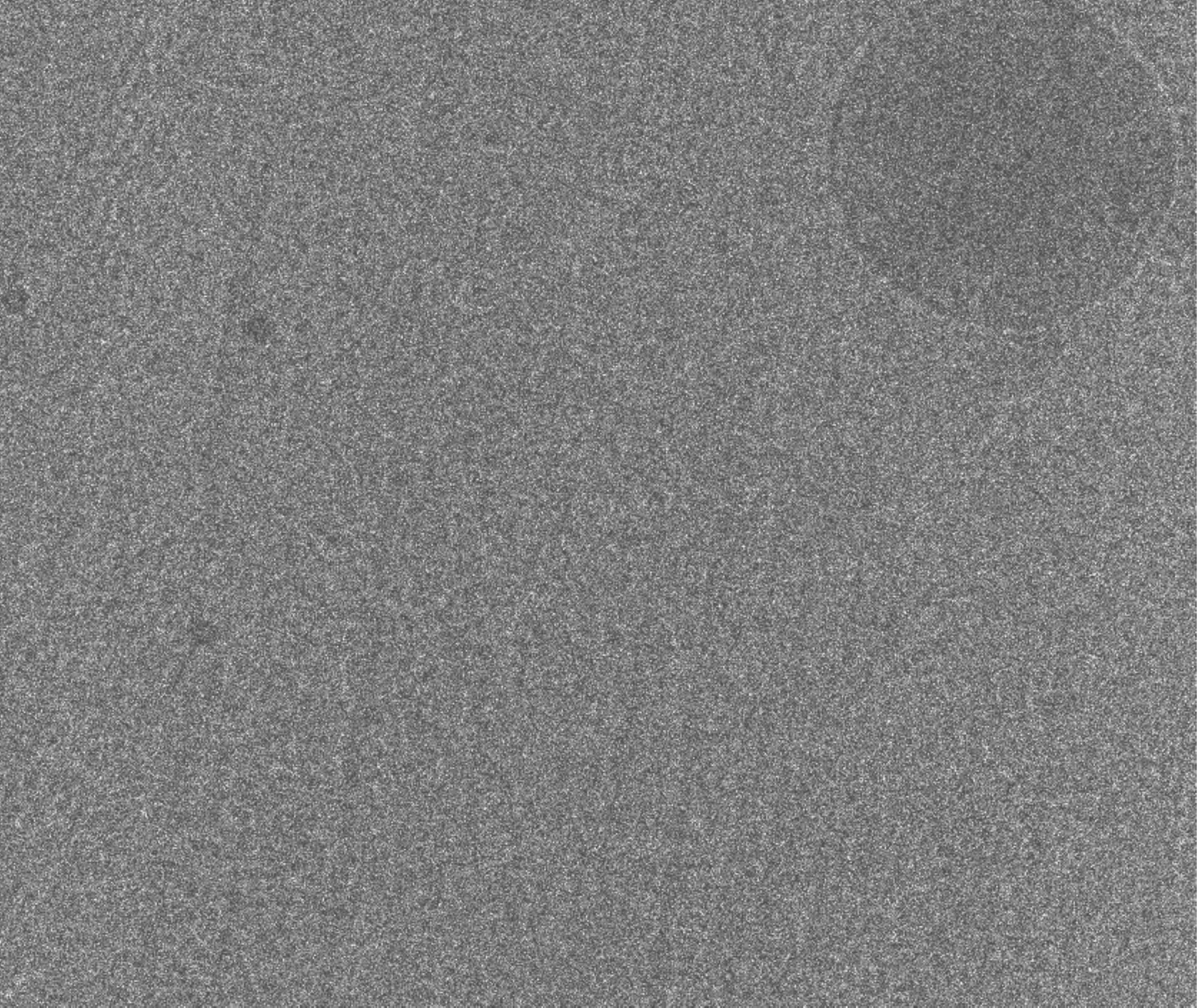}
        \put (59,68) {\figNum{red}{{\ding{228}}}}
        \put (63.5,25.9) {\figNum{yellow}{\rotatebox{170}{{\ding{228}}}}}
        \put (51.5,22.0) {\figNum{yellow}{\rotatebox{350}{{\ding{228}}}}}
        \put (18,35.4) {\figNum{blue}{\rotatebox{180}{\ding{228}}}}
    \end{overpic}}};
\node[anchor=south west,inner sep=0] at (5.68,0) {
    {\begin{overpic}[width=2.74cm]{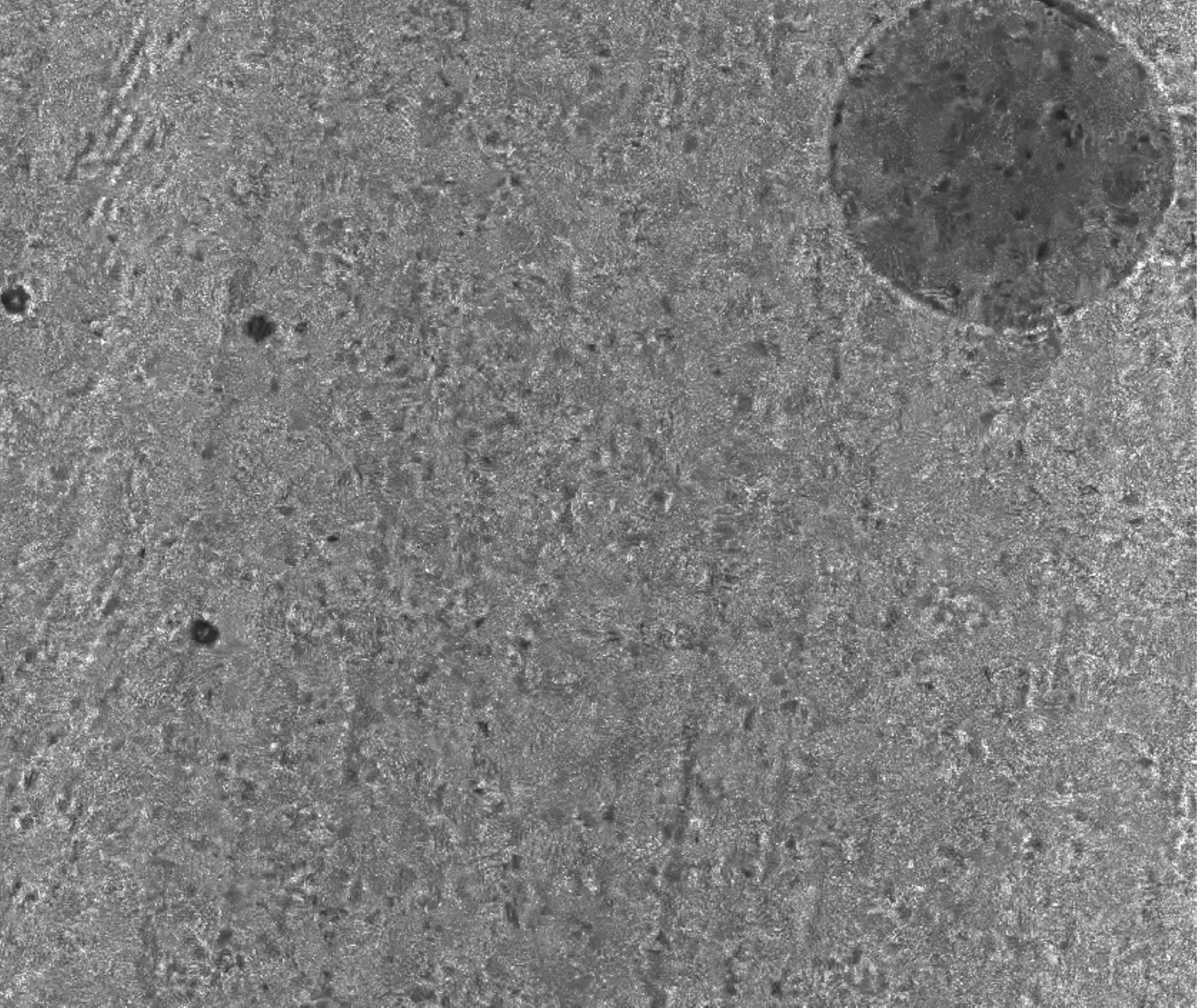}
        \put (25, 2) {\contour{black}{\figOverText{white}{Runtime: \texttildelow 33.2s}}}
        \put (59,68) {\figNum{red}{{\ding{228}}}}
        \put (63.5,25.9) {\figNum{yellow}{\rotatebox{170}{{\ding{228}}}}}
        \put (51.5,22.0) {\figNum{yellow}{\rotatebox{350}{{\ding{228}}}}}
        \put (18,35.4) {\figNum{blue}{\rotatebox{180}{\ding{228}}}}
    \end{overpic}}};
\node[anchor=south west,inner sep=0] at (8.52,0) {
    {\begin{overpic}[width=2.74cm]{figs/results_placeholder_white_frame}
        \put (29, 34) {\scalebox{4}{$\varnothing$}}
        \put (-1, 16) {\makebox[80pt]{\Centerstack{\scriptsize{Clean target}}}}
        \put (-1, 6) {\makebox[80pt]{\Centerstack{\scriptsize{not available.}}}}
    \end{overpic}}};
\node[anchor=south west,inner sep=0] at (11.36,0) {
    {\begin{overpic}[width=2.74cm]{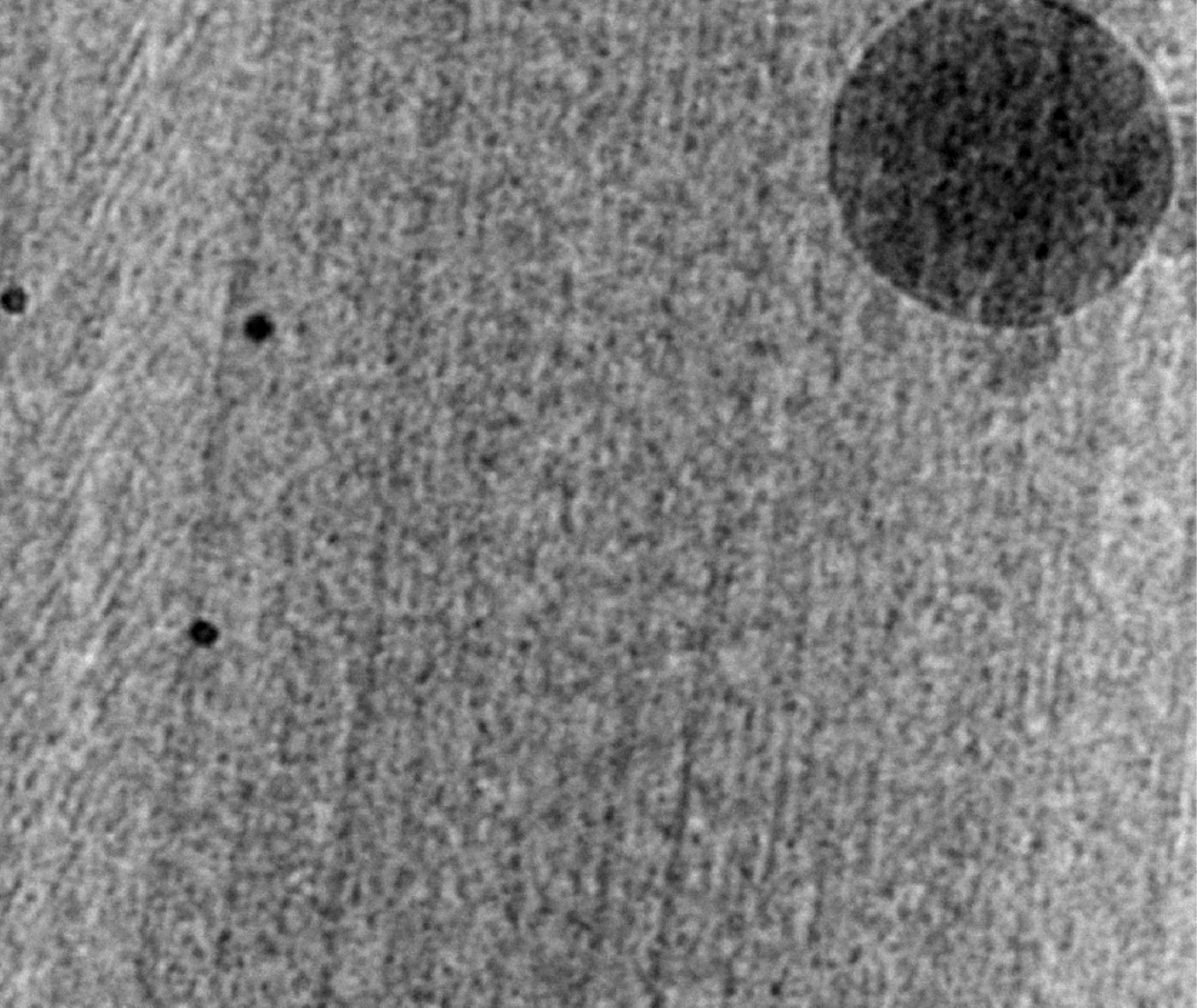}
        \put (30, 2) {\contour{black}{\figOverText{white}{Runtime: \texttildelow 1.3s}}}
        \put (59,68) {\figNum{red}{{\ding{228}}}}
        \put (63.5,25.9) {\figNum{yellow}{\rotatebox{170}{{\ding{228}}}}}
        \put (51.5,22.0) {\figNum{yellow}{\rotatebox{350}{{\ding{228}}}}}
        \put (18,35.4) {\figNum{blue}{\rotatebox{180}{\ding{228}}}}
    \end{overpic}}};
\node[anchor=south west,inner sep=0] at (14.20,0) {
    {\begin{overpic}[width=2.74cm]{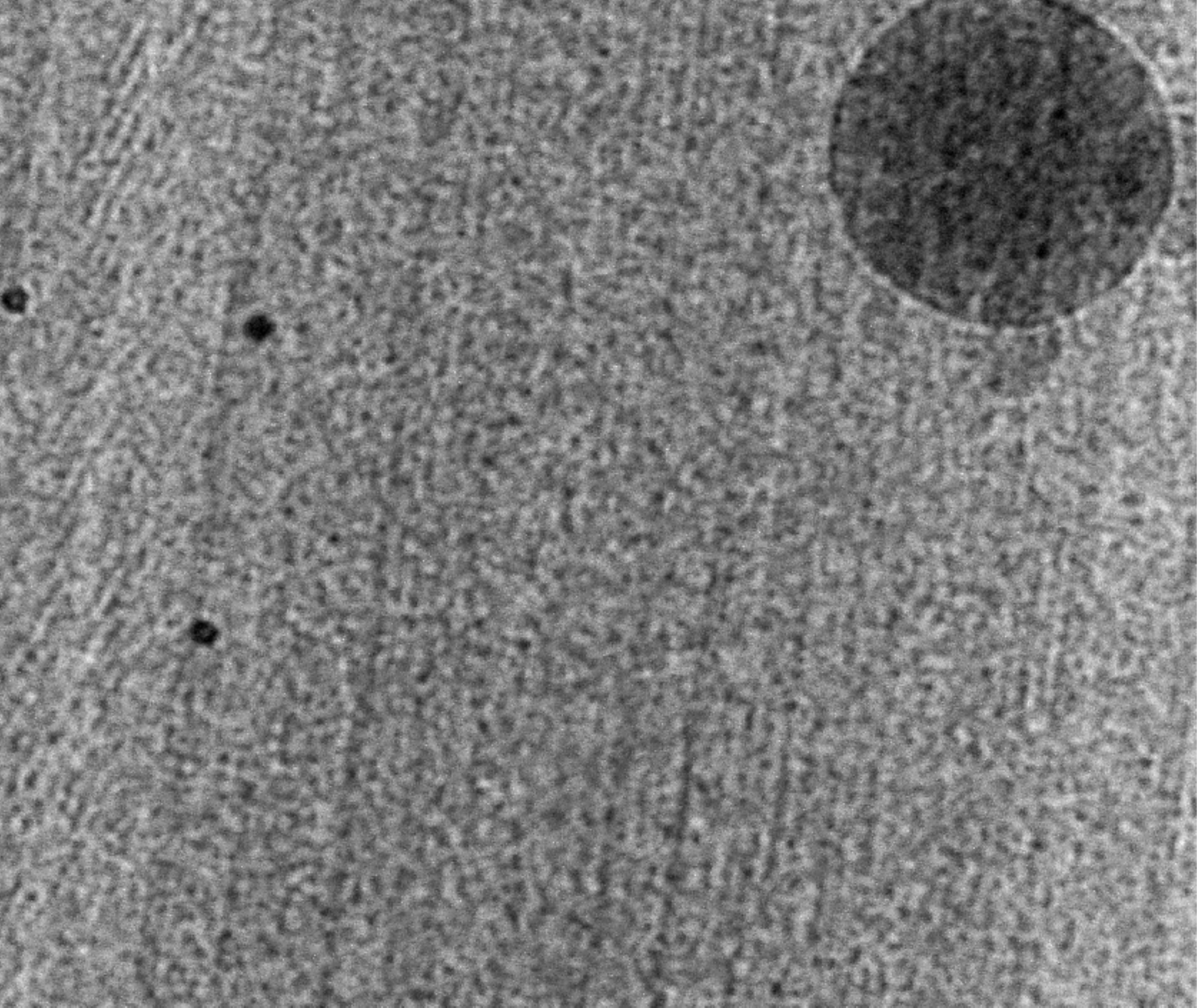}
        \put (30, 2) {\contour{black}{\figOverText{white}{Runtime: \texttildelow 1.3s}}}
        \put (59,68) {\figNum{red}{{\ding{228}}}}
        \put (63.5,25.9) {\figNum{yellow}{\rotatebox{170}{{\ding{228}}}}}
        \put (51.5,22.0) {\figNum{yellow}{\rotatebox{350}{{\ding{228}}}}}
        \put (18,35.4) {\figNum{blue}{\rotatebox{180}{\ding{228}}}}
    \end{overpic}}};
\end{tikzpicture}
\end{minipage}
}

\vspace{0.6mm}
\hspace{3.6mm}
\centerline{
\begin{minipage}{\linewidth}
\begin{tikzpicture}
\node[anchor=south west,inner sep=0] at (0,0) {
    {\begin{overpic}[width=2.74cm]{figs/results_placeholder_white_frame}
        \put (37, 34) {\scalebox{4}{?}}
        \put (-1, 16) {\makebox[80pt]{\Centerstack{\scriptsize{Does not exist.}}}}
        \put (-12, 14) {\rotatebox{90}{\small\bf CTC-MSC}}
    \end{overpic}}};
\node[anchor=south west,inner sep=0] at (2.84,0) {
    {\begin{overpic}[width=2.74cm]{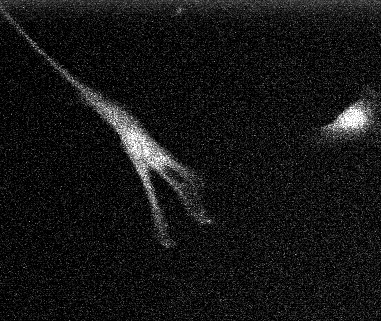}
    \end{overpic}}};
\node[anchor=south west,inner sep=0] at (5.68,0) {
    {\begin{overpic}[width=2.74cm]{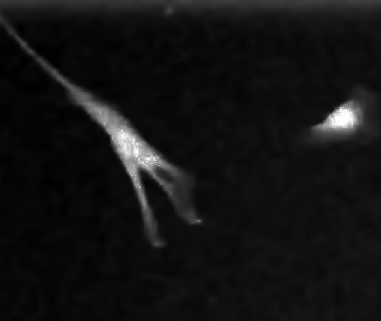}
        \put (30, 2) {\contour{black}{\figOverText{white}{Runtime: \texttildelow 4.6s}}}
    \end{overpic}}};
\node[anchor=south west,inner sep=0] at (8.52,0) {
    {\begin{overpic}[width=2.74cm]{figs/results_placeholder_white_frame}
        \put (29, 34) {\scalebox{4}{$\varnothing$}}
        \put (-1, 16) {\makebox[80pt]{\Centerstack{\scriptsize{Clean target}}}}
        \put (-1, 6) {\makebox[80pt]{\Centerstack{\scriptsize{not available.}}}}
    \end{overpic}}};
\node[anchor=south west,inner sep=0] at (11.36,0) {
    {\begin{overpic}[width=2.74cm]{figs/results_placeholder_white_frame}
        \put (29, 34) {\scalebox{4}{$\varnothing$}}
        \put (-1, 16) {\makebox[80pt]{\Centerstack{\scriptsize{Noisy target}}}}
        \put (-1, 6) {\makebox[80pt]{\Centerstack{\scriptsize{not available.}}}}
    \end{overpic}}};
\node[anchor=south west,inner sep=0] at (14.20,0) {
    {\begin{overpic}[width=2.74cm]{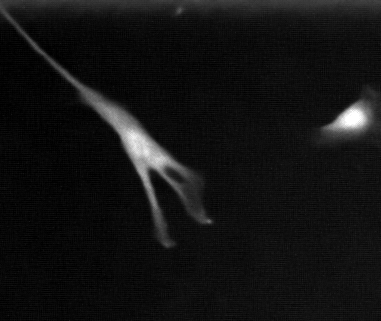}
        \put (30, 2) {\contour{black}{\figOverText{white}{Runtime: \texttildelow 0.1s}}}
    \end{overpic}}};
\end{tikzpicture}
\end{minipage}
}

\vspace{0.6mm}
\hspace{3.6mm}
\centerline{
\begin{minipage}{\linewidth}
\begin{tikzpicture}
\node[anchor=south west,inner sep=0] at (0,0) {
    {\begin{overpic}[width=2.74cm]{figs/results_placeholder_white_frame}
        \put (37, 34) {\scalebox{4}{?}}
        \put (-1, 16) {\makebox[80pt]{\Centerstack{\scriptsize{Does not exist.}}}}
        \put (-12, 13) {\rotatebox{90}{\small\bf CTC-N2DH}}
    \end{overpic}}};
\node[anchor=south west,inner sep=0] at (2.84,0) {
    {\begin{overpic}[width=2.74cm]{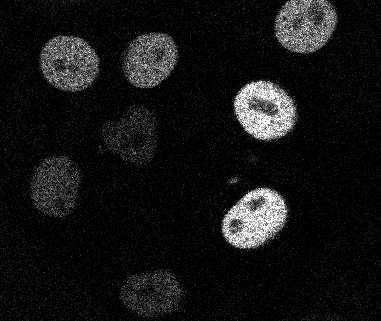}
    \end{overpic}}};
\node[anchor=south west,inner sep=0] at (5.68,0) {
    {\begin{overpic}[width=2.74cm]{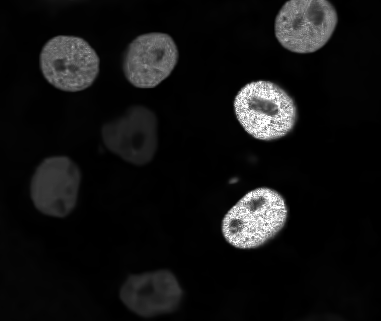}
        \put (30, 2) {\contour{black}{\figOverText{white}{Runtime: \texttildelow 5.2s}}}
    \end{overpic}}};
\node[anchor=south west,inner sep=0] at (8.52,0) {
    {\begin{overpic}[width=2.74cm]{figs/results_placeholder_white_frame}
        \put (29, 34) {\scalebox{4}{$\varnothing$}}
        \put (-1, 16) {\makebox[80pt]{\Centerstack{\scriptsize{Clean target}}}}
        \put (-1, 6) {\makebox[80pt]{\Centerstack{\scriptsize{not available.}}}}
    \end{overpic}}};
\node[anchor=south west,inner sep=0] at (11.36,0) {
    {\begin{overpic}[width=2.74cm]{figs/results_placeholder_white_frame}
        \put (29, 34) {\scalebox{4}{$\varnothing$}}
        \put (-1, 16) {\makebox[80pt]{\Centerstack{\scriptsize{Noisy target}}}}
        \put (-1, 6) {\makebox[80pt]{\Centerstack{\scriptsize{not available.}}}}
    \end{overpic}}};
\node[anchor=south west,inner sep=0] at (14.20,0) {
    {\begin{overpic}[width=2.74cm]{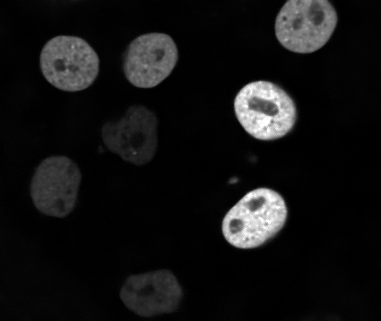}
        \put (30, 2) {\contour{black}{\figOverText{white}{Runtime: \texttildelow 0.1s}}}
    \end{overpic}}};
\end{tikzpicture}
\end{minipage}
}
\caption{Results and average PSNR values obtained by BM3D, traditionally trained, \NtoN trained, and \NtoV trained denoising networks.
For BSD68 data and simulated data all methods are applicable.
For cryo-TEM data ground truth images are unobtainable.
Since pairs of noisy images are available, we can still perform \NoiseNoise training.
Red, yellow, and blue arrowheads indicate an ice artifact, two tubulin protofilaments that are known to be $4nm$ apart, and a $10nm$ gold bead, respectively.
For the CTC-MSC and CTC-N2DH data only single noisy images exist.
Hence, neither traditional nor \NtoN training is applicable, while our proposed training scheme can still be applied.
}
\label{fig:n2vResults}
\vspace{-2mm}
\end{figure*}
}

\newcommand\figTeaser{
\begin{figure}[t]
\centerline{
\begin{minipage}{.33\linewidth}
    \begin{minipage}{\linewidth}
    {\begin{overpic}[width=1.305cm]{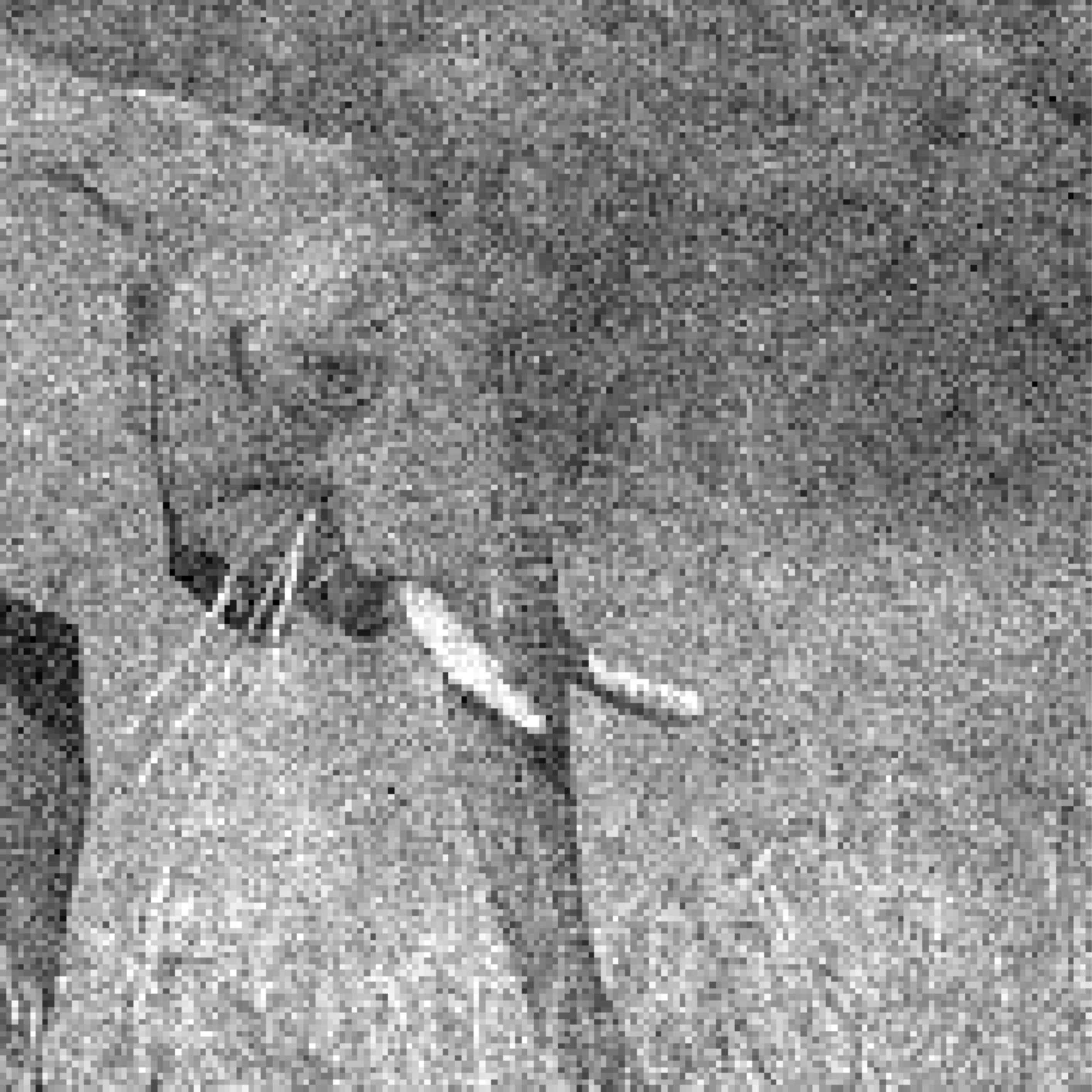}
        \put (3, 82) {\figOverText{white}{noisy}}
    \end{overpic}}
    {\begin{overpic}[width=1.305cm]{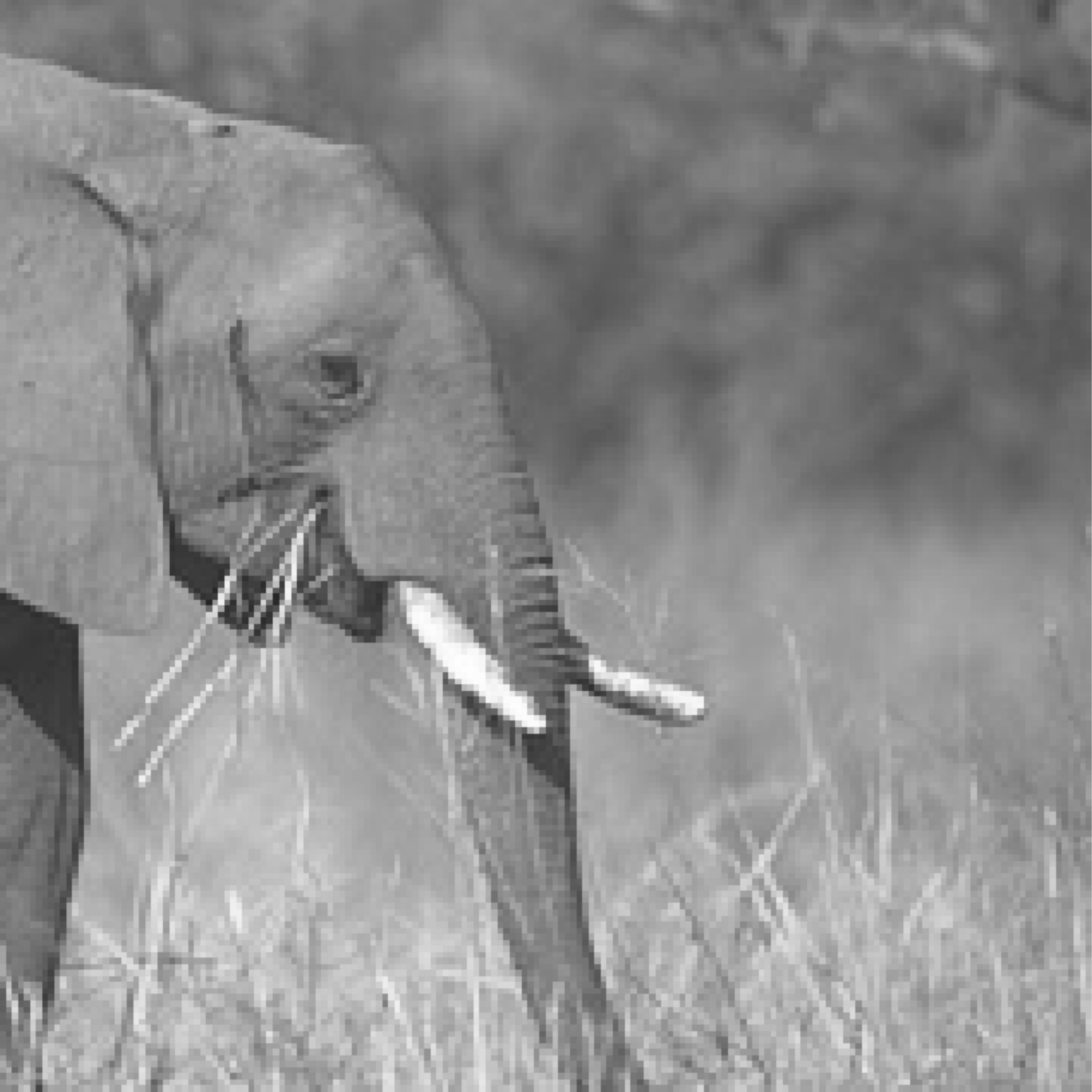}
        \put (3, 82) {\figOverText{white}{clean}}
    \end{overpic}}\\\vspace{-3.7mm}
    \end{minipage}
    {\begin{overpic}[width=2.70cm]{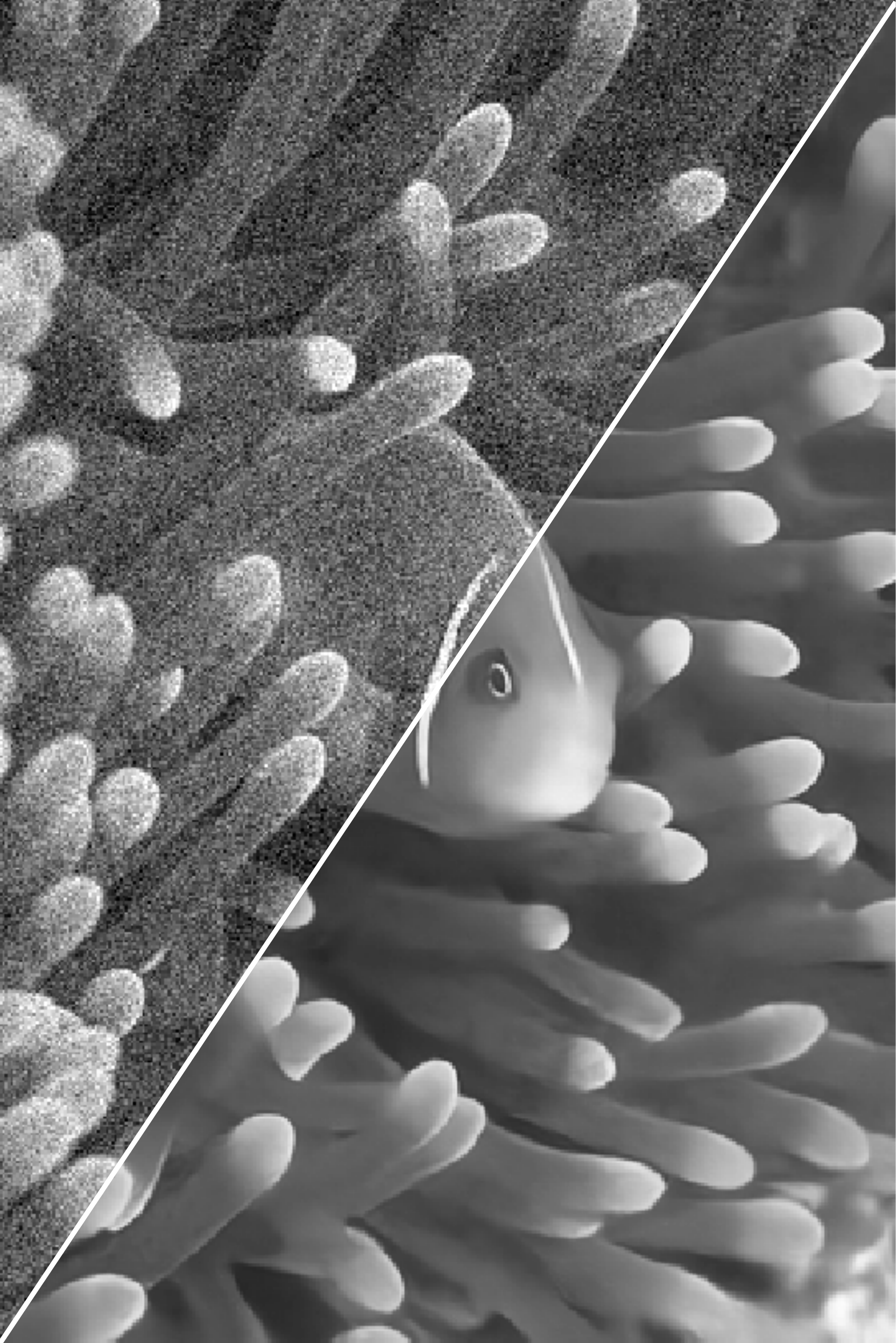}
     \put (29.5, 1.5) {\contour{black}{\figOverText{white}{Traditional}}}
     \put (1.2, 94.5) {\contour{black}{\figOverText{white}{Input}}}
    \end{overpic}}
\end{minipage}
\begin{minipage}{.33\linewidth}
    \begin{minipage}{\linewidth}
    {\begin{overpic}[width=1.305cm]{figs/n2v_teaser_train_noisy0}
        \put (3, 82) {\figOverText{white}{noisy}}
    \end{overpic}}
    {\begin{overpic}[width=1.305cm]{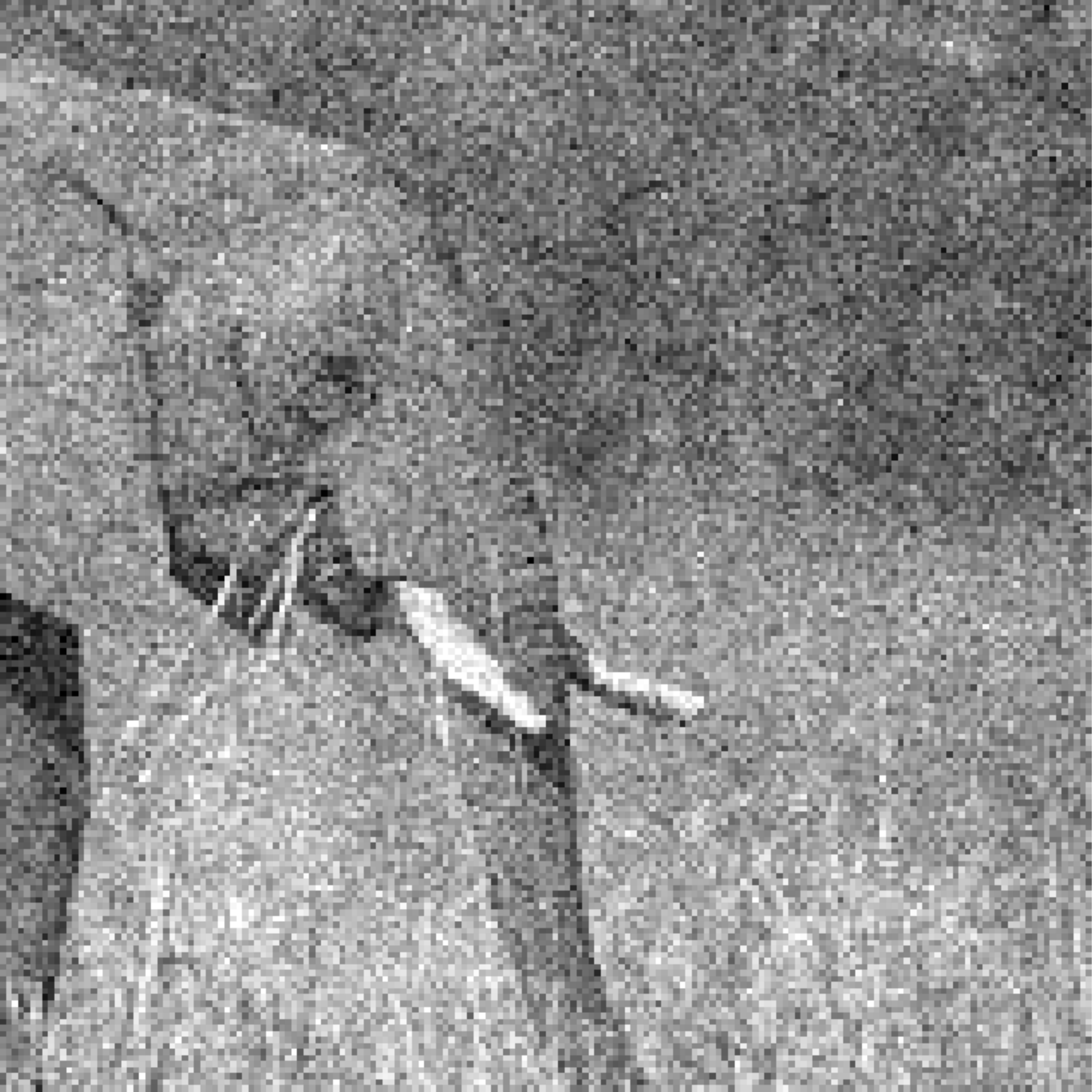}
        \put (3, 82) {\figOverText{white}{noisy}}
    \end{overpic}}\\\vspace{-3.7mm}
    \end{minipage}
    {\begin{overpic}[width=2.70cm]{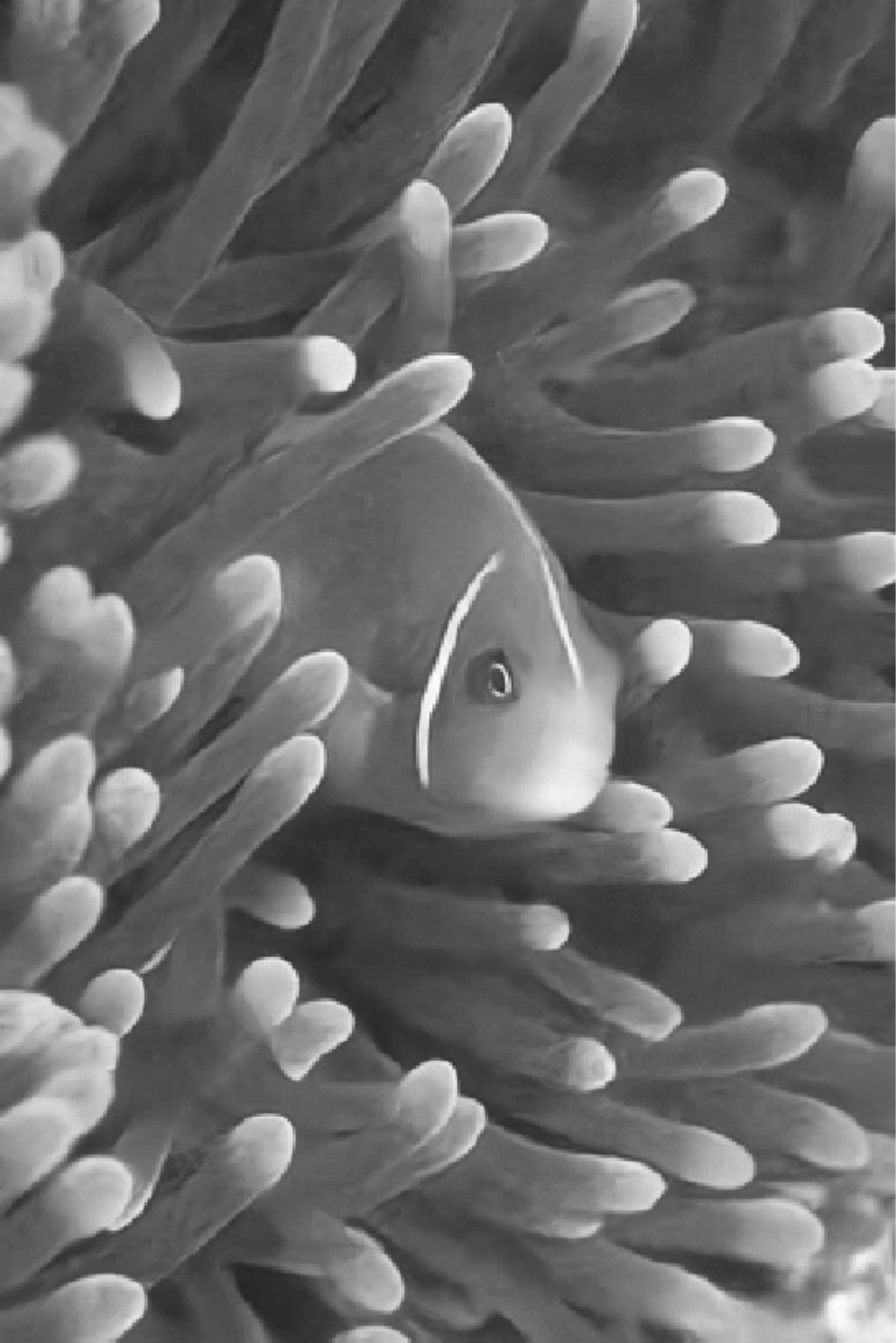}
     \put (22.5, 1.5) {\contour{black}{\figOverText{white}{\NoiseNoise}}}
    \end{overpic}}
\end{minipage}
\begin{minipage}{.33\linewidth}
\begin{minipage}{\linewidth}
    {\begin{overpic}[width=1.305cm]{figs/n2v_teaser_train_noisy0}
        \put (3, 82) {\figOverText{white}{noisy}}
    \end{overpic}}
    {\begin{overpic}[width=1.305cm]{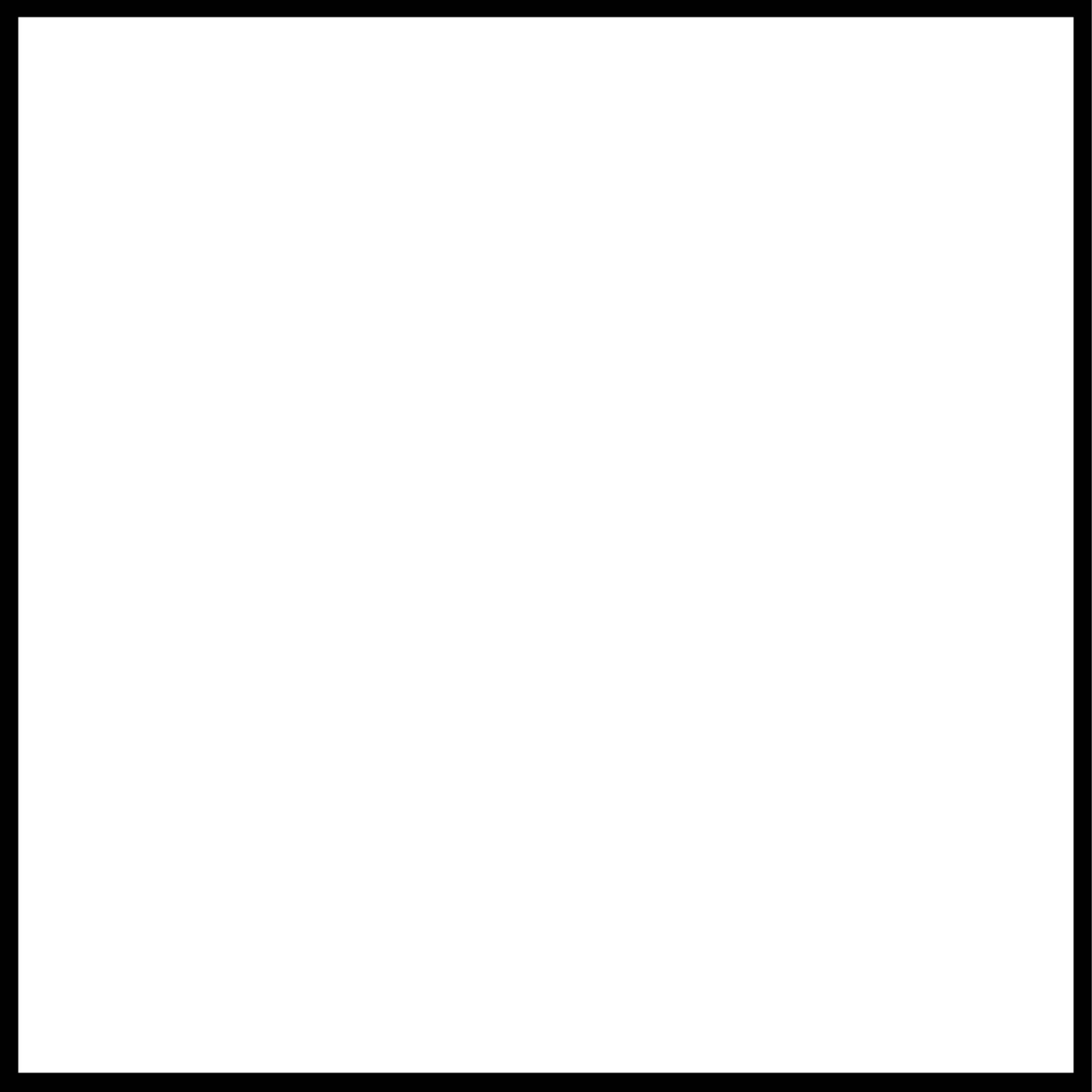}
        \put (3, 82) {\figOverText{black}{void}}
    \end{overpic}}\\\vspace{-3.7mm}
    \end{minipage}
    {\begin{overpic}[width=2.70cm]{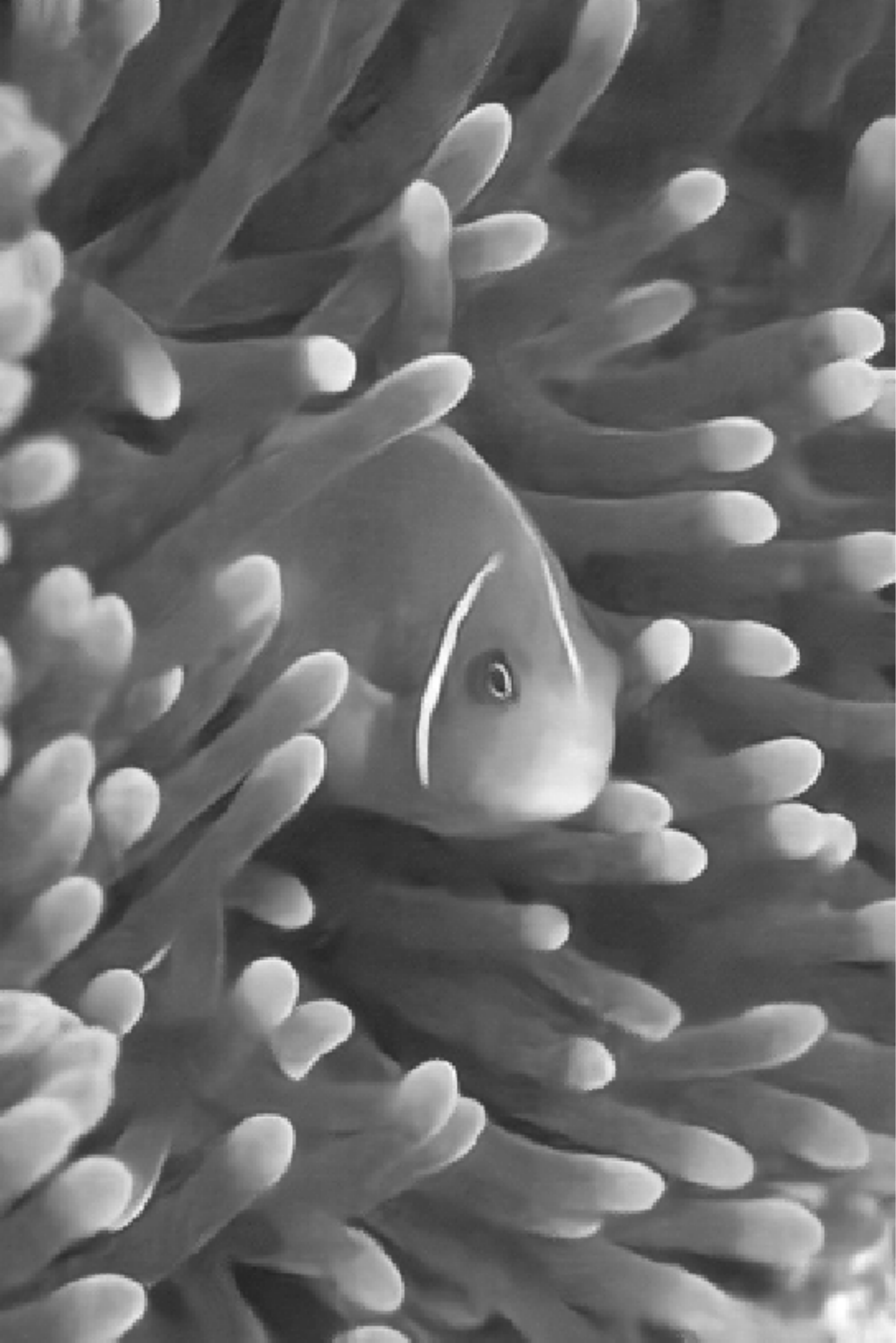}
     \put (20.5, 1.5) {\contour{black}{\textcolor{white}{\small{\bf \NoiseVoid}}}}
    \end{overpic}}
\end{minipage}
}
\caption{
Training schemes for CNN-based denoising.
Traditionally, training networks for denoising requires pairs of noisy and clean images.
For many practical applications, however, clean target images are not available.
\NoiseNoise (\NtoN)~\cite{noise2noise} enables the training of CNNs from independent pairs of noisy images.
Still, also noisy image pairs are not usually available.
This motivated us to propose \NoiseVoid~(\NtoV), a novel training procedure that does not require noisy image pairs, nor clean target images.
By enabling CNNs to be trained directly on a body of noisy images, we open the door to a plethora of new applications, \eg on biomedical data.
}
\label{fig:n2vTeaser}
\vspace{-2mm}
\end{figure}
}

\newcommand\figLimitationRare{
\begin{figure}[htb]
\centerline{
\begin{minipage}{.33\linewidth}
\begin{overpic}[width=2.7cm]{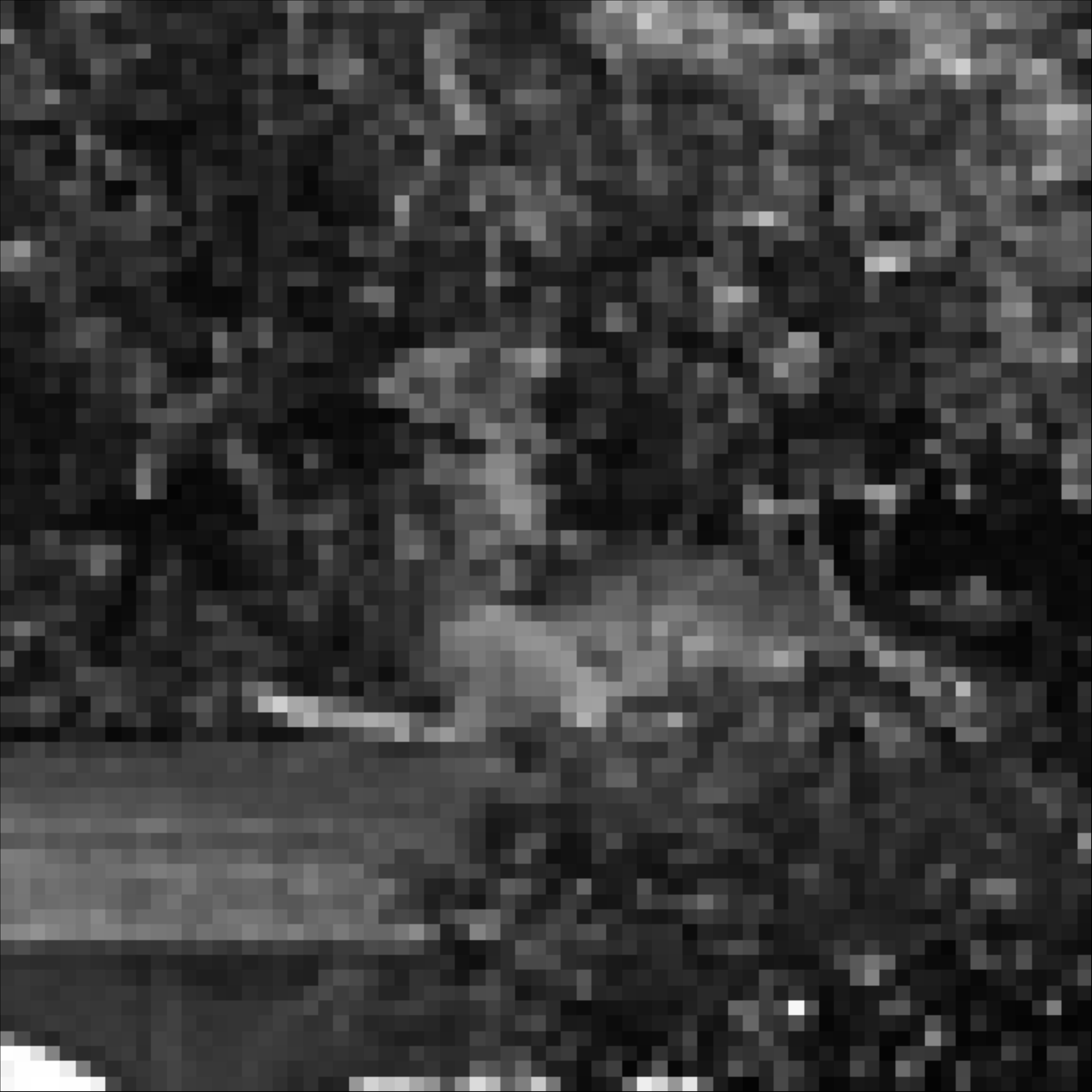}
    \put (1.5, 2.5) {\figNum{white}{(a)}}
    \put (62.4,12.5) {\rotatebox{315}{\figNum{red}{\bf\ding{228}}}}
\end{overpic}
\end{minipage}
\begin{minipage}{.33\linewidth}
\begin{overpic}[width=2.7cm]{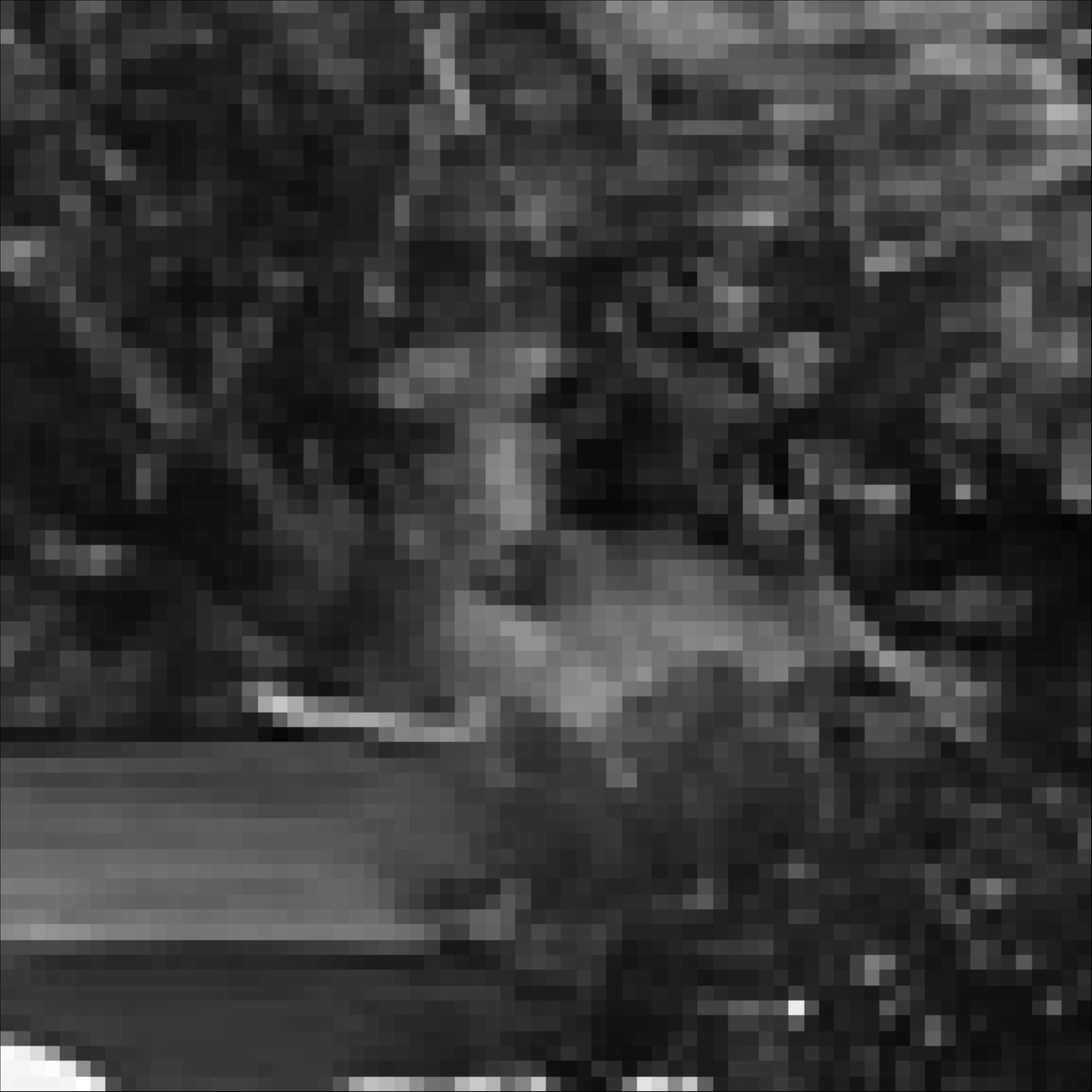}
    \put (1.5, 2.5) {\figNum{white}{(b)}}
    \put (62.4,12.5) {\rotatebox{315}{\figNum{red}{\bf\ding{228}}}}
\end{overpic}
\end{minipage}
\begin{minipage}{.33\linewidth}
\begin{overpic}[width=2.7cm]{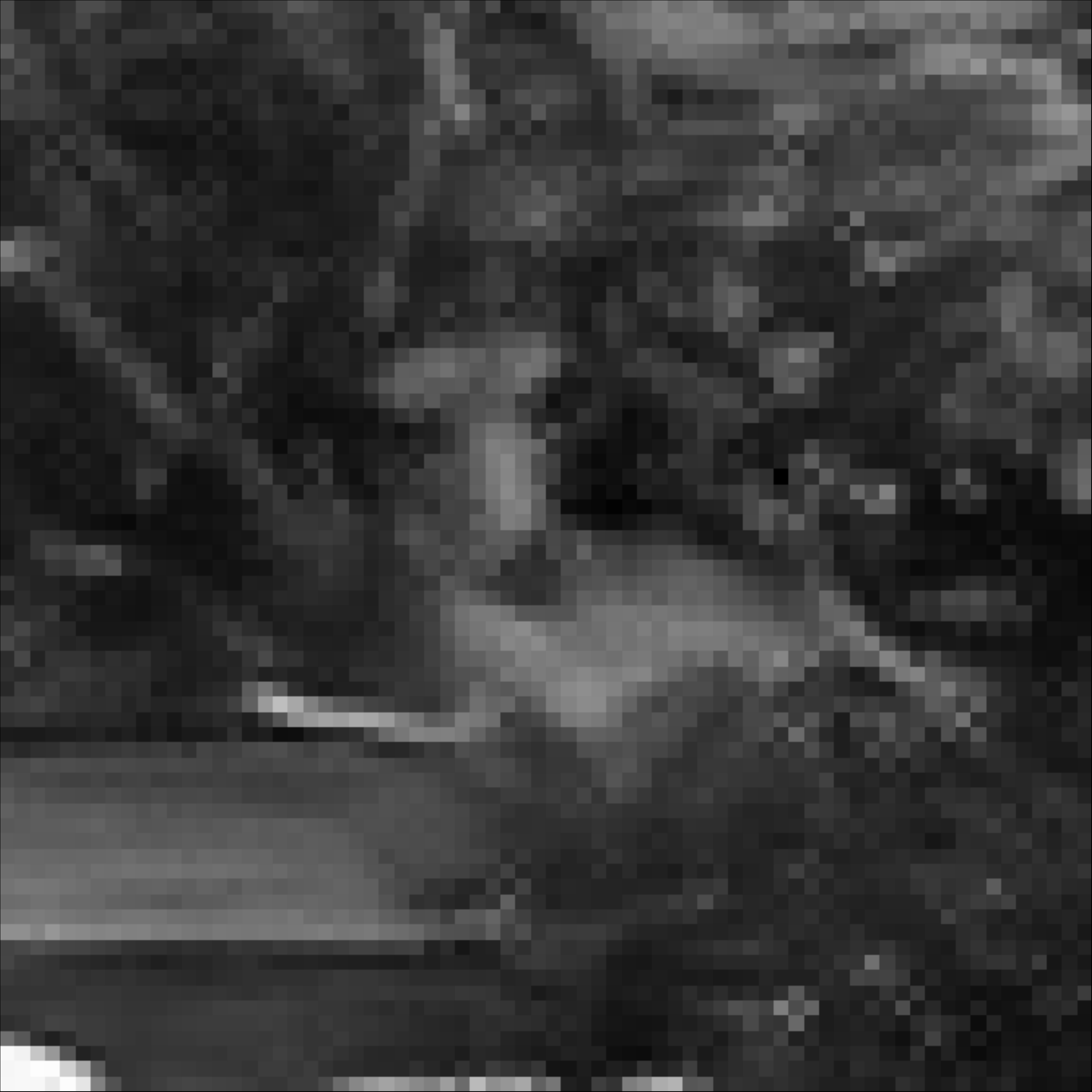}
    \put (1.5, 2.5) {\figNum{white}{(c)}}
    \put (62.4,12.5) {\rotatebox{315}{\figNum{red}{\bf\ding{228}}}}
\end{overpic}
\end{minipage}
}\vspace{2pt}
\centerline{
\begin{minipage}{.33\linewidth}
\begin{overpic}[width=2.7cm]{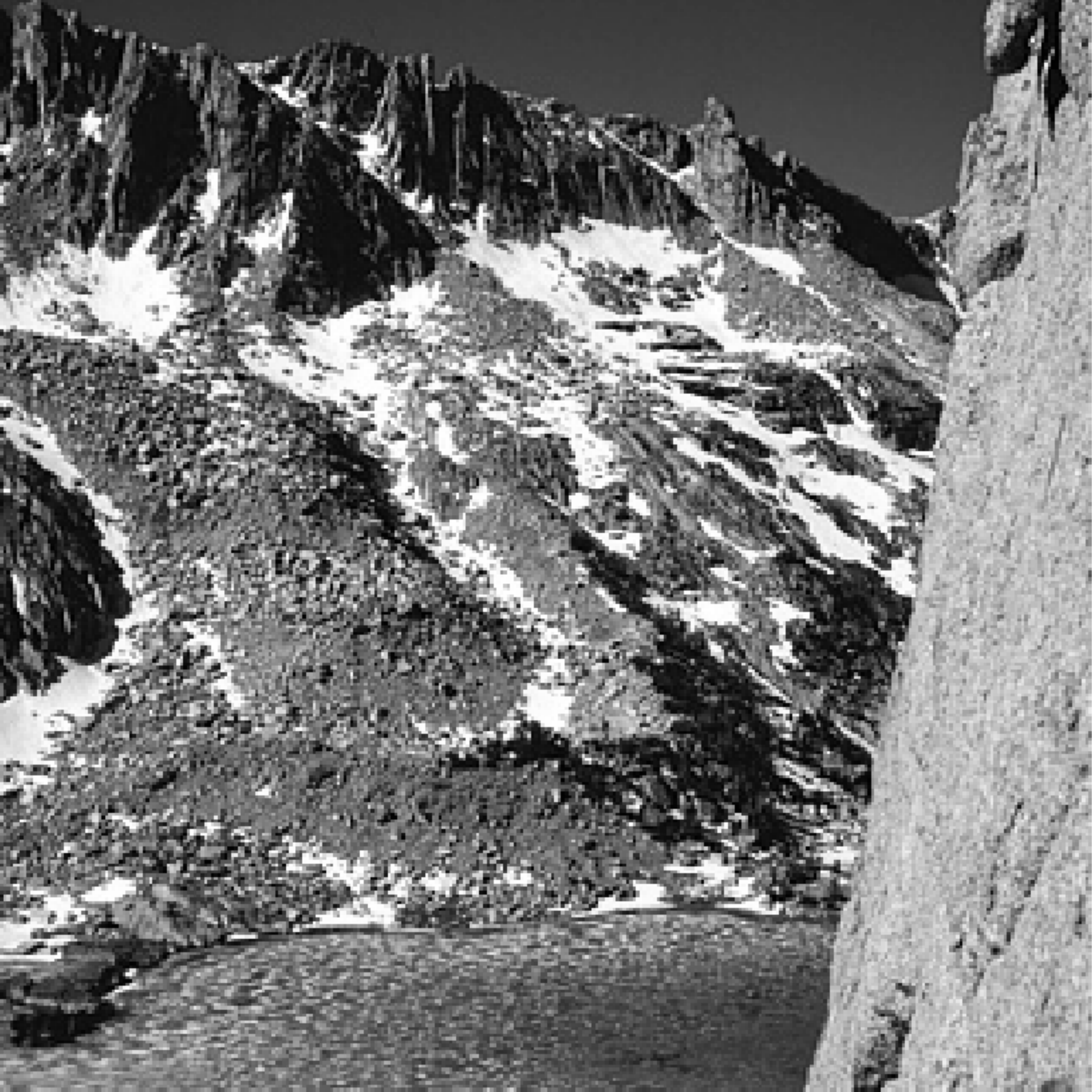}
    \put (1.5, 2.5) {\figNum{white}{(d)}}
\end{overpic}
\end{minipage}
\begin{minipage}{.33\linewidth}
\begin{overpic}[width=2.7cm]{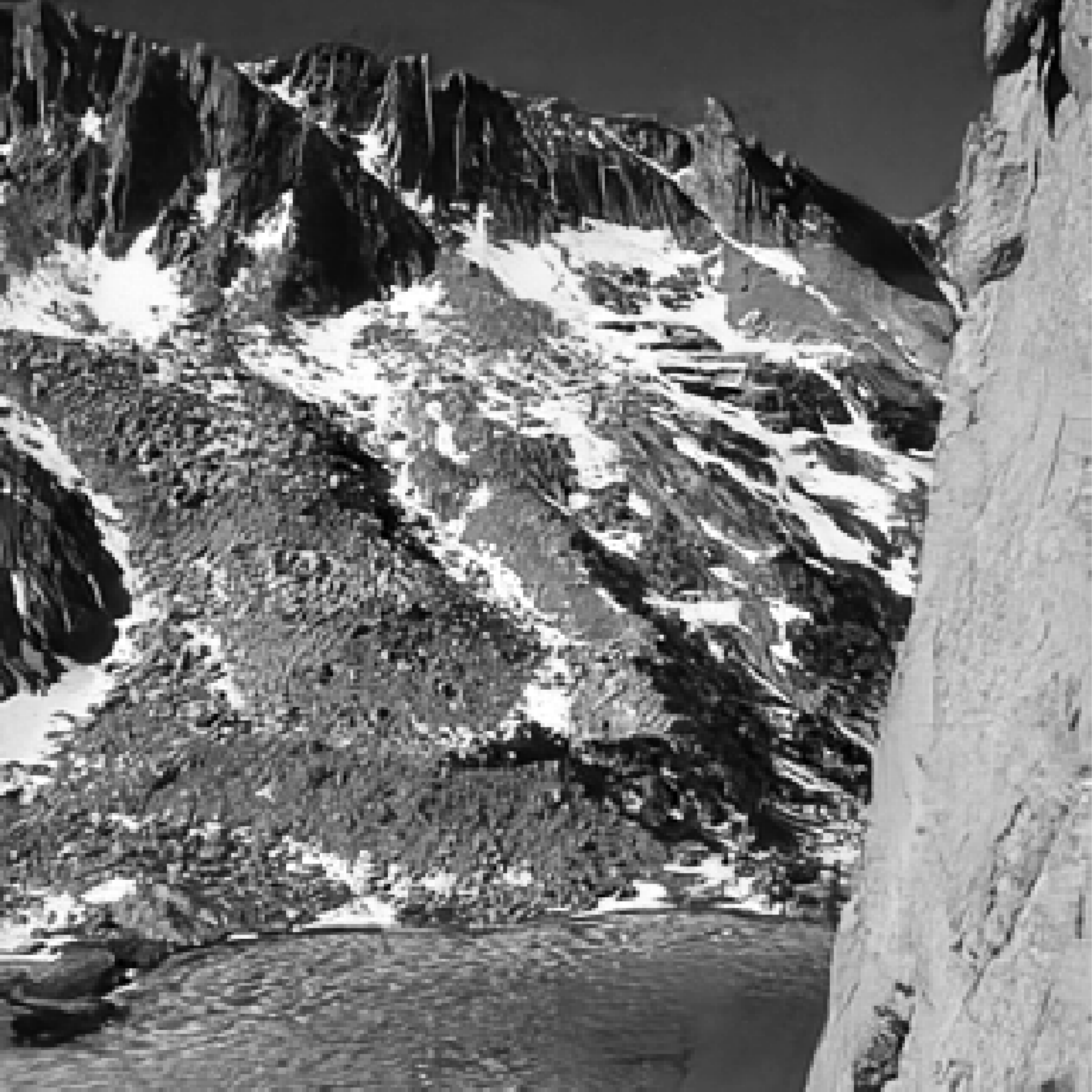}
    \put (1.5, 2.5) {\figNum{white}{(e)}}
\end{overpic}
\end{minipage}
\begin{minipage}{.33\linewidth}
\begin{overpic}[width=2.7cm]{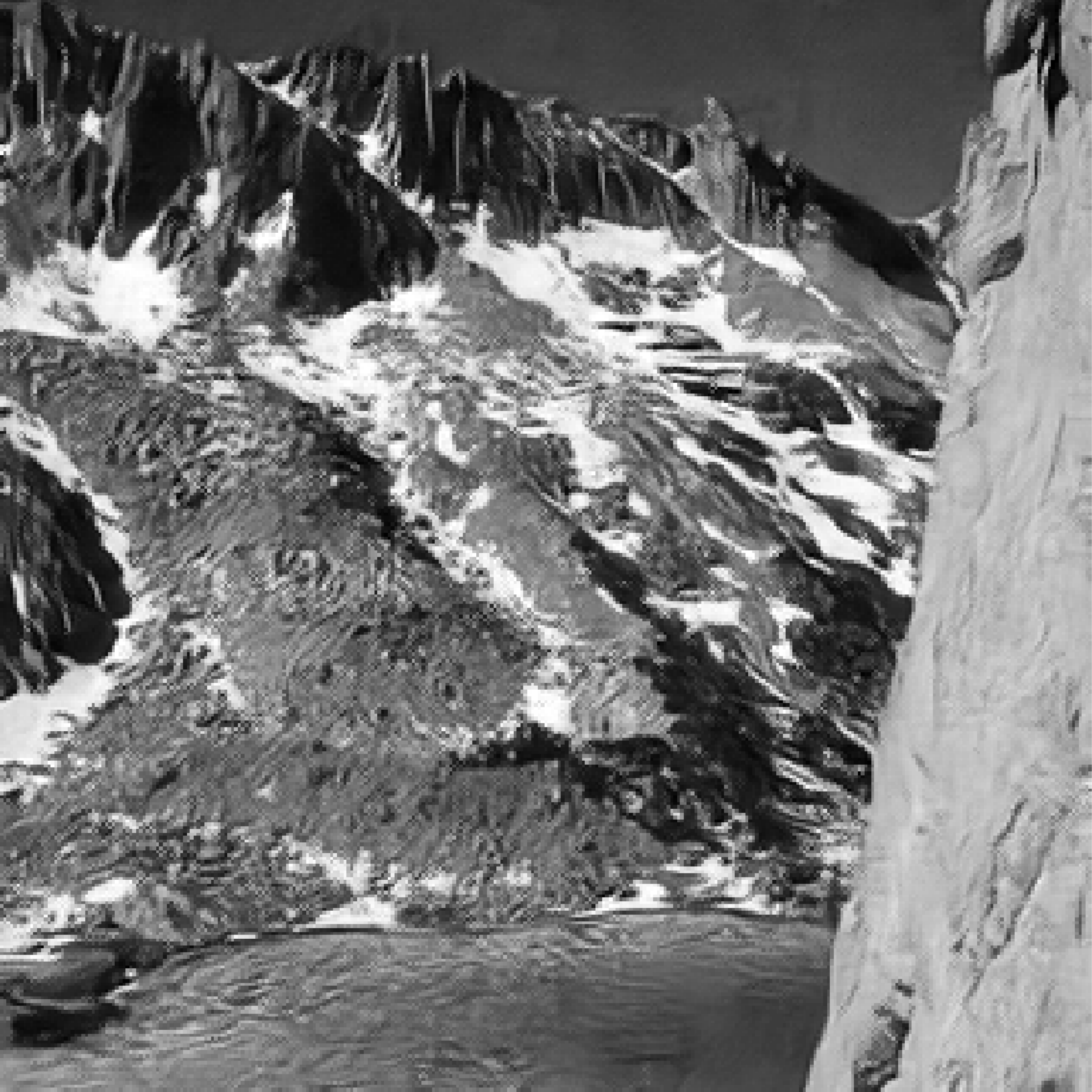}
    \put (1.5, 2.5) {\figNum{white}{(f)}}
\end{overpic}
\end{minipage}
}
\caption{
Failure cases of \NtoV trained networks.
{\bf (a)}~A crop from the ground truth test image with the largest individual pixel error (indicated by red arrow).
{\bf (b)}~Result of a traditionally trained network on the same image.
{\bf (c)}~Result of our \NtoV trained network.
The network fails to predict this bright and isolated pixel.
{\bf (d)}~A crop from the ground truth test image with the largest total error.
{\bf (e)}~Result of a traditionally trained network on the same image.
{\bf (f)}~Result of our \NtoV trained network.
Both networks are not able to preserve the grainy structure of the image, but the \NtoV trained network loses more high-frequency detail.
}
\label{fig:limitationRare}
\vspace{-2mm}
\end{figure}
}

\newcommand\figLimitationPattern{
\begin{figure}[htb]
\centerline{
\begin{minipage}{\linewidth}
\begin{minipage}{.32\linewidth}
\begin{overpic}[width=2.7cm]{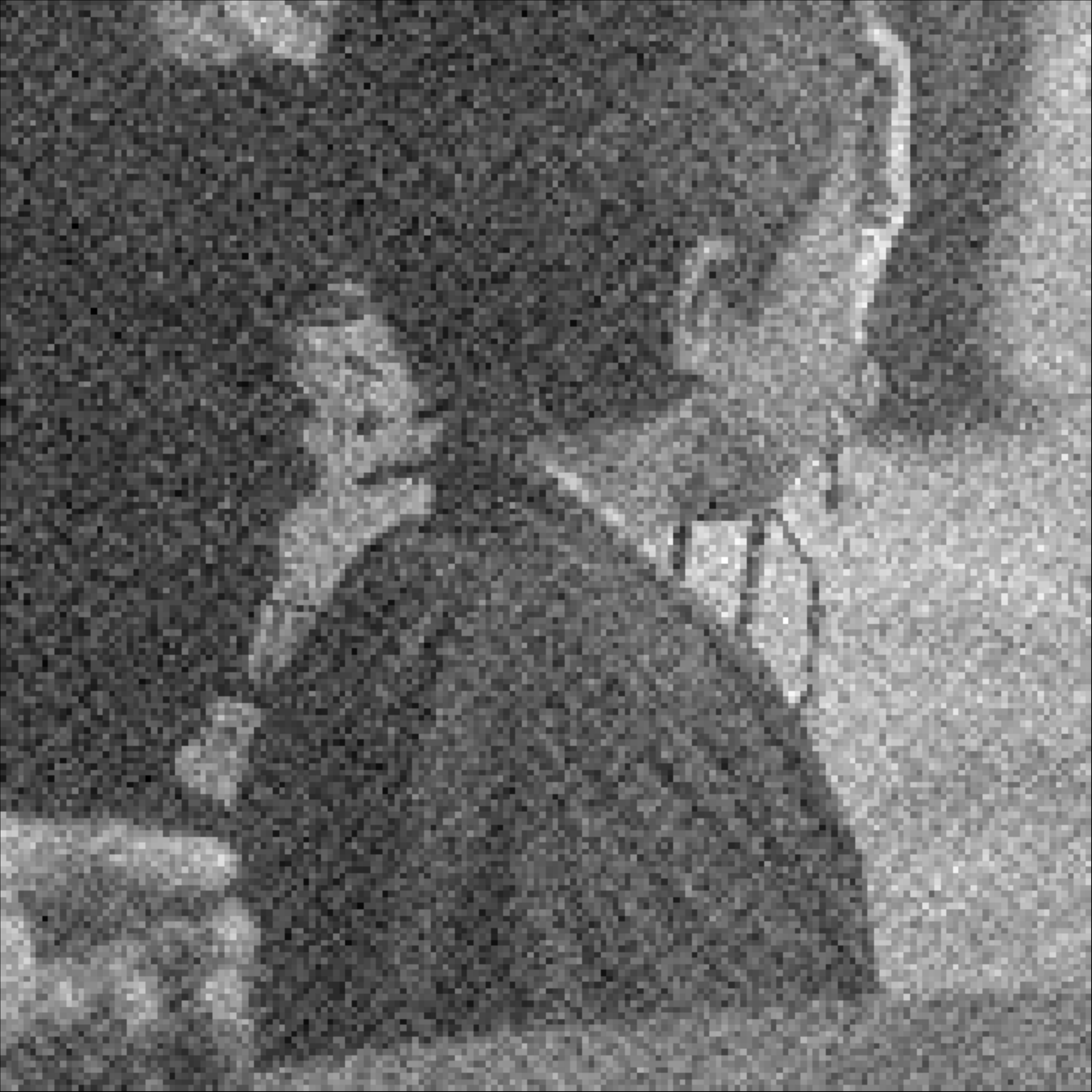}
    \put (1.5, 2.5) {\figNum{white}{(a)}}
\end{overpic}
\end{minipage}
\begin{minipage}{.32\linewidth}
\begin{overpic}[width=2.7cm]{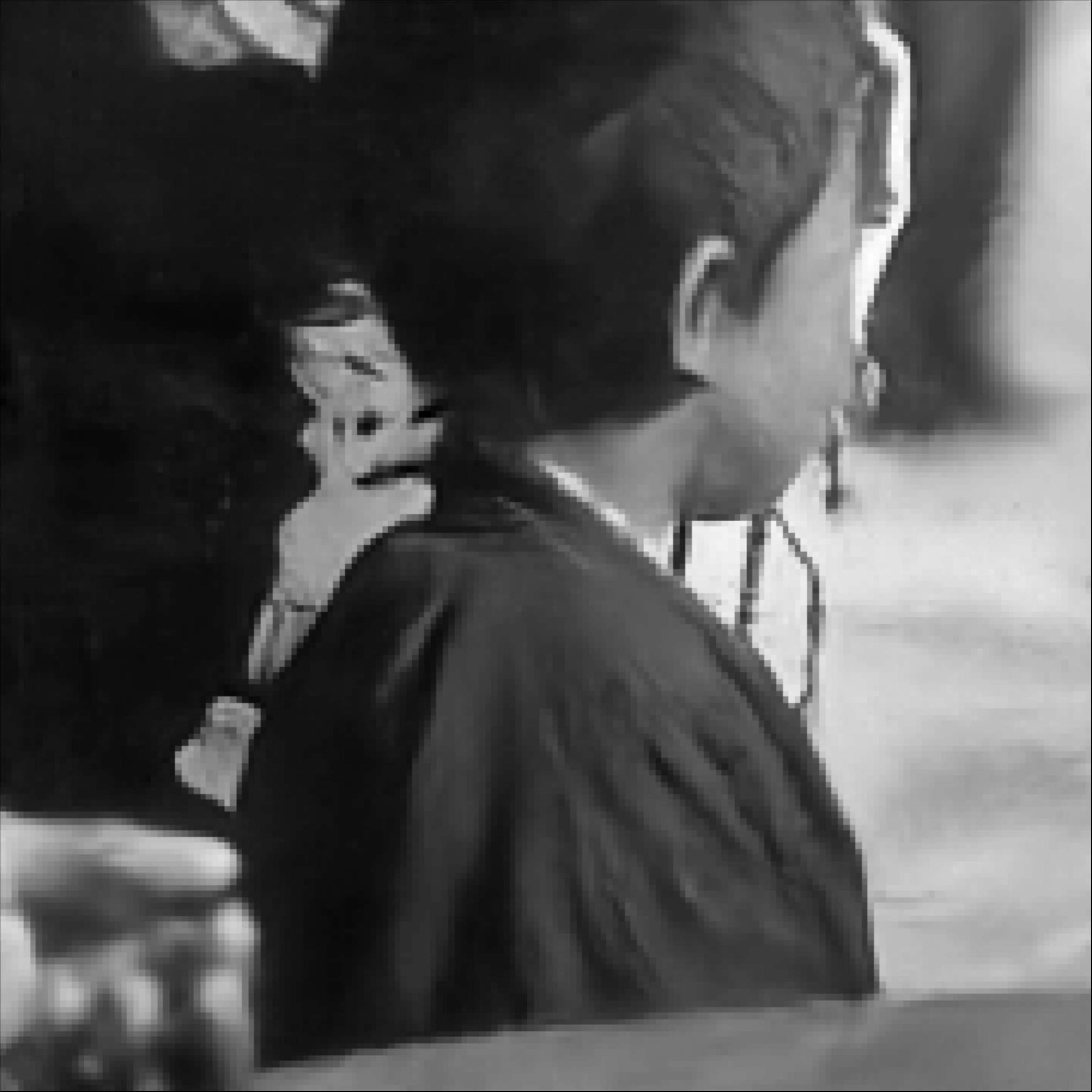}
    \put (1.5, 2.5) {\figNum{white}{(b)}}
\end{overpic}
\end{minipage}
\begin{minipage}{.32\linewidth}
\begin{overpic}[width=2.7cm]{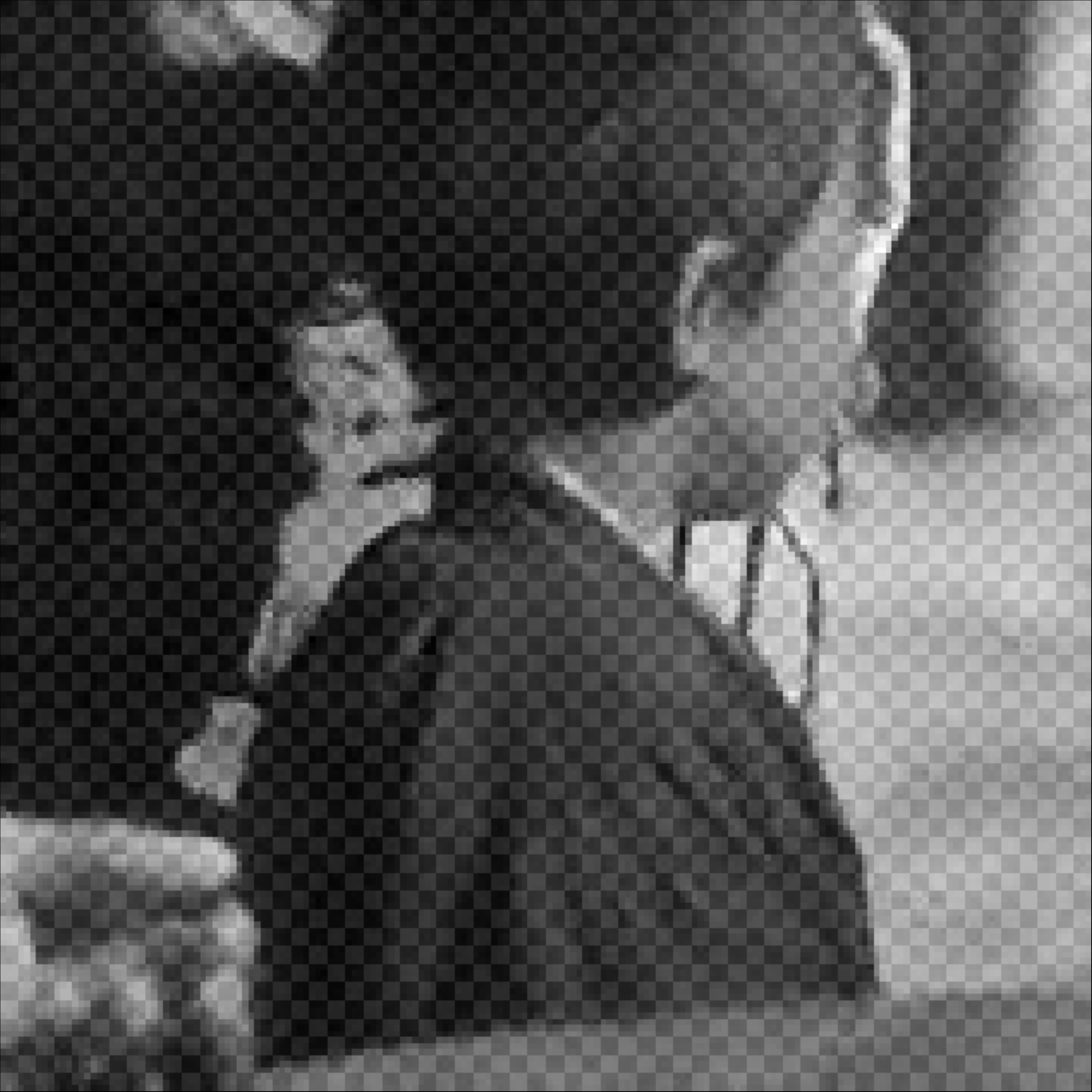}
    \put (1.5, 2.5) {\figNum{white}{(c)}}
\end{overpic}
\end{minipage}
\end{minipage}
}
\vspace{1pt}
\centerline{
\begin{minipage}{\linewidth}
\begin{minipage}{.32\linewidth}
\begin{overpic}[width=2.7cm]{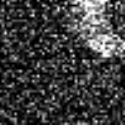}
    \put (1.5, 2.5) {\figNum{white}{(d)}}
\end{overpic}
\end{minipage}
\begin{minipage}{.32\linewidth}
\vspace{1pt}
\begin{overpic}[width=2.7cm]{figs/place_holder_white_frame}
        \put (29, 42) {\scalebox{4}{$\varnothing$}}
        \put (-2, 24) {\makebox[80pt]{\Centerstack{\scriptsize{Clean target}}}}
        \put (-2, 14) {\makebox[80pt]{\Centerstack{\scriptsize{not available.}}}}
\end{overpic}
\end{minipage}
\begin{minipage}{.32\linewidth}
\begin{overpic}[width=2.7cm]{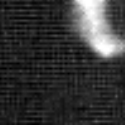}
    \put (1.5, 2.5) {\figNum{white}{(e)}}
\end{overpic}
\end{minipage}
\end{minipage}
}
\caption{
Effect of structured noise on \NtoV trained network predictions.
Structured noise violates our assumption that noise is pixel-independent (see also Eq.~\ref{eq:noise}).
{\bf (a)}~A photograph corrupted by structured noise.
The hidden checkerboard pattern is barely visible.
{\bf (b)}~The denoised result of a traditionally trained CNN.
{\bf (c)}~The denoised result of an \NtoV trained CNN.
The independent components of the noise are removed, but the structured components remain.
{\bf (d)}~Structured noise in real microscopy data.
{\bf (e)}~The denoised result of an \NtoV trained CNN.
A hidden pattern in the noise is revealed.
Note that due to the lacking training data, it is not possible to use \NtoN or the traditional training scheme in this case.
}
\label{fig:limitationPattern}
\vspace{-2mm}
\end{figure}
}